\documentclass{ieeeaccess}

\usepackage{times}
\usepackage[T1]{fontenc}
\usepackage[latin9]{inputenc}
\usepackage{color}
\usepackage{array}
\usepackage{tikz}
\NewSpotColorSpace{PANTONE}
\AddSpotColor{PANTONE} {PANTONE3015C} {PANTONE\SpotSpace 3015\SpotSpace C} {1 0.3 0 0.2}
\SetPageColorSpace{PANTONE}%
\usepackage{amsmath}
\usepackage{amssymb}
\usepackage{graphicx}
\PassOptionsToPackage{normalem}{ulem}
\usepackage{ulem}
\usepackage{nameref}
\makeatletter
\usepackage[english]{babel}
\providecommand{\tabularnewline}{\\}
\usepackage[T1]{fontenc}
\usepackage{array}
\usepackage{cite} %
\usepackage{color}

\usepackage{enumitem}

\usepackage{listings}
\usepackage{caption}
\usepackage{graphics}
\usepackage{placeins}
\usepackage{graphicx, epstopdf}

\usepackage{colortbl}

\usepackage{multirow}

\usepackage{floatrow}
\usepackage[label font=bf,labelformat=simple, position=top]{subfig}%
\definecolor{celadon}{rgb}{0.67, 0.88, 0.69}
\definecolor{hellgelb}{rgb}{1,1,0.85} 
\definecolor{colKeys}{rgb}{0,0,1} 
\definecolor{colIdentifier}{rgb}{0,0,0} 
\definecolor{colComments}{rgb}{0,0.5,0} 
\definecolor{colString}{rgb}{0.81,0.12,0.95}
\definecolor{deepblue}{rgb}{0,0,1}
\definecolor{deepred}{rgb}{0.6,0,0}
\definecolor{deepgreen}{rgb}{0,0.5,0}
\definecolor{blue_light}{RGB}{16,161,239}

\usepackage{hyperref}
\hypersetup{
  colorlinks,
  citecolor=black,
  linkcolor=black,
  urlcolor=blue}
\usepackage[algo2e,ruled,vlined,linesnumbered]{algorithm2e}
\usepackage{subfig}
\usepackage{caption}

\usepackage{boldline} 
\usepackage{empheq} 
\usepackage{textcomp} 
\usepackage{tablefootnote}
\usepackage{newfloat}
\DeclareFloatingEnvironment[fileext=lop]{movie}
\makeatother

\newcommand{\addaccess}[1]{#1}

\newcommand{\addrevision}[1]{#1}
\newcommand{\addrevisiongraphics}[1]{#1}

\usepackage{cite}
\usepackage{amsmath,amssymb,amsfonts}
\usepackage{algorithmic}
\usepackage{graphicx}
\def\BibTeX{{\rm B\kern-.05em{\sc i\kern-.025em b}\kern-.08em
		T\kern-.1667em\lower.7ex\hbox{E}\kern-.125emX}}
\begin{document}
	\history{Date of publication xxxx 00, 0000, date of current version xxxx 00, 0000.}
	\doi{10.1109/ACCESS.2017.DOI}
	
	\title{PANTHER: Perception-Aware Trajectory Planner in Dynamic Environments}
	\author{\uppercase{Jesus Tordesillas}\authorrefmark{1}, \IEEEmembership{Student Member, IEEE} and
		\uppercase{Jonathan P. How}\authorrefmark{1}, \IEEEmembership{Fellow, IEEE}}
	\address[1]{Aerospace Controls Laboratory, Massachusetts Institute of Technology, Cambridge, MA 02139 USA (e-mail: jtorde@mit.edu, jhow@mit.edu)}
	\tfootnote{This work was supported in part by Boeing Research
		\& Technology}
	
	\markboth
	{Author \headeretal: Preparation of Papers for IEEE TRANSACTIONS and JOURNALS}
	{Author \headeretal: Preparation of Papers for IEEE TRANSACTIONS and JOURNALS}
	
	\corresp{Corresponding author: Jesus Tordesillas (e-mail: jtorde@mit.edu).}

\begin{abstract}
This paper presents PANTHER, a real-time perception-aware~(PA) trajectory planner \addaccess{for multirotor-UAVs (Unmanned Aerial Vehicles)} in dynamic environments. PANTHER plans trajectories that avoid dynamic obstacles while also keeping them in the sensor field of view~(FOV) and minimizing the blur to aid in object tracking. The rotation and translation of the UAV are jointly optimized, which allows PANTHER to fully exploit the differential flatness of multirotors to maximize the PA objective. Real-time performance is achieved by implicitly imposing the \addaccess{underactuated dynamics} of the UAV through the Hopf fibration. PANTHER is able to keep the obstacles inside the FOV~\addaccess{7.9 and~1.5} times more than non-PA approaches and PA approaches that decouple translation and yaw, respectively. The projected velocity (and hence the blur) is reduced by~\addaccess{18\% and~34\%}, respectively. \addaccess{This leads to average success rates three times larger than state-of-the-art approaches} in multi-obstacle avoidance scenarios. The MINVO basis is used to impose low-conservative collision avoidance constraints in position and velocity space. Finally, extensive hardware experiments in unknown dynamic environments with all the computation running onboard are presented, with velocities of up to 5.8~m/s, and with relative velocities (with respect to the obstacles) of up to 6.3~m/s. The only sensors used are an IMU, a forward-facing depth camera, and a downward-facing monocular camera. 
\end{abstract}


\titlepgskip=-15pt

\maketitle

\noindent
\textbf{Video}: \href{https://youtu.be/jKmyW6v73tY}{https://youtu.be/jKmyW6v73tY} \\
\textbf{Code}: \href{https://github.com/mit-acl/panther}{https://github.com/mit-acl/panther}

\section{Introduction and Related Work}\label{sec:introduction}

\definecolor{color_formulation}{RGB}{179,227,204}
\definecolor{color_goal}{RGB}{245,247,167}

\newcommand*\circled[5]{\tikz[baseline=-3]{
		\node[shape=circle, fill=#2, draw=#3, text=#4, inner sep=0.8pt, scale=#5] (char) {\textbf{#1}};}}

\newcommand{\tikzrectangle}[2][black,fill=red]{\tikz[baseline=0.0ex, line width=0.2mm]\draw[#1] [#1] (0,0) rectangle (0.2,0.2);}%

\newcommand{\tikzcircle}[2][black,fill=red]{\tikz[baseline=0.0ex, line width=0.3mm]\draw[#1] [#1] (0,0.08) circle (0.08);}%

\newcommand{\velunits}{m/s}
\newcommand{\accelunits}{m/s\textsuperscript{2}}
\newcommand{\jerkunits}{m/s\textsuperscript{3}}

\newcommand{\circleform}[1]{\circled{A.#1}{color_formulation!50}{black}{black}{0.6}}%
\newcommand{\circlegoal}[1]{\circled{B.#1}{color_goal!50}{black}{black}{0.6}}%

\newcommand{\citenoPA}{\cite{chen2016online,gao2020teach,tordesillas2019faster,lin2020robust,falanga2020dynamic,wang2021autonomous,sanket2020evdodgenet}}
\newcommand{\citePAHw}{\cite{nageli2017real,bonatti2018autonomous,chen2020bio,ding2019efficient}}
\newcommand{\citePADec}{\cite{zhou2020raptor,spasojevic2020perception,murali2019perception}}
\newcommand{\citePAJoint}{\cite{watterson2020trajectory,falanga2018pampc,penin2018vision}}
\newcommand{\citeReduceStateEst}{\cite{spasojevic2020perception,murali2019perception,falanga2018pampc,watterson2020trajectory,spasojevic2020joint,bartolomei2020perception,lee2020aggressive,zhang2018perception,costante2016perception,achtelik2014motion,penin2017vision,preiss2018simultaneous,frey2019towards,salaris2019online}}
\newcommand{\citeRecordTarget}{\cite{thomas2017autonomous,jeon2020detection,penin2018vision,guanrui2021pcmpc,penin2017vision,chen2017using,jeon2019online}}

\begin{table*}
	\noindent\resizebox{\columnwidth}{!}{%
	\begin{centering}
	\caption{Classification of the related work, together with a (nonexhaustive) list of references. \label{tab:classification} }		\begin{tabular}{|c|c|c|}
			\hline 
			\multicolumn{2}{|c|}{\cellcolor{color_formulation!50} \textbf{Formulation}} & \textbf{Ref.} \tabularnewline
			\hline 
			\hline 
			\multicolumn{2}{|l|}{\circleform{1} \textbf{Not PA}} & \citenoPA{} \tabularnewline
			\hline 
			\multicolumn{2}{|l|}{\circleform{2} \textbf{PA with additional hardware}} & \citePAHw{}\tabularnewline
			\hline 
			\multirow{2}{*}{\circleform{3} \textbf{PA planning}} & \textbf{Decoupling}  & \citePADec{} \tabularnewline
			\cline{2-3} \cline{3-3} 
			& \textbf{Joint opt.} & \citePAJoint{}, PANTHER \tabularnewline
			\hline 
		\end{tabular}%
	\hspace{0.8cm}
		\begin{tabular}{|l|c|}
			\hline 
			\multicolumn{1}{|c|}{\cellcolor{color_goal!50}\textbf{Goal}} & \textbf{Ref.}\tabularnewline
			\hline 
			\hline 
			\circlegoal{1} \textbf{Reduce state estimation uncertainty} & \citeReduceStateEst{} \tabularnewline
			\hline 
			\circlegoal{2} \textbf{Record/chase a target} & \citeRecordTarget{} \tabularnewline
			\hline 
			\circlegoal{3} \textbf{Avoidance of dynamic obstacles} & PANTHER\tabularnewline
			\hline 
		\end{tabular}
		\par\end{centering}
}
\end{table*}

\PARstart{W}{hile} \addaccess{the last decade has seen an increase on the number of successful deployments of multirotor-UAVs in different real-world scenarios, their applicability is often limited by two common assumptions, namely the fact that the environment is static, and/or the omnidirectional coverage of the sensor(s) of the UAV.  Indeed, many UAVs have a limited 
 FOV, and many applications (delivery, aerial videography, emergency response, etc.) have non-static environments due to the presence of cars, people, and/or other UAVs.  
 Hence, relaxing these assumptions is critical to fully exploit the potential of the UAVs and expand the range of their possible applications.}

\begin{figure}
	\begin{centering}
		\includegraphics[width=1\columnwidth]{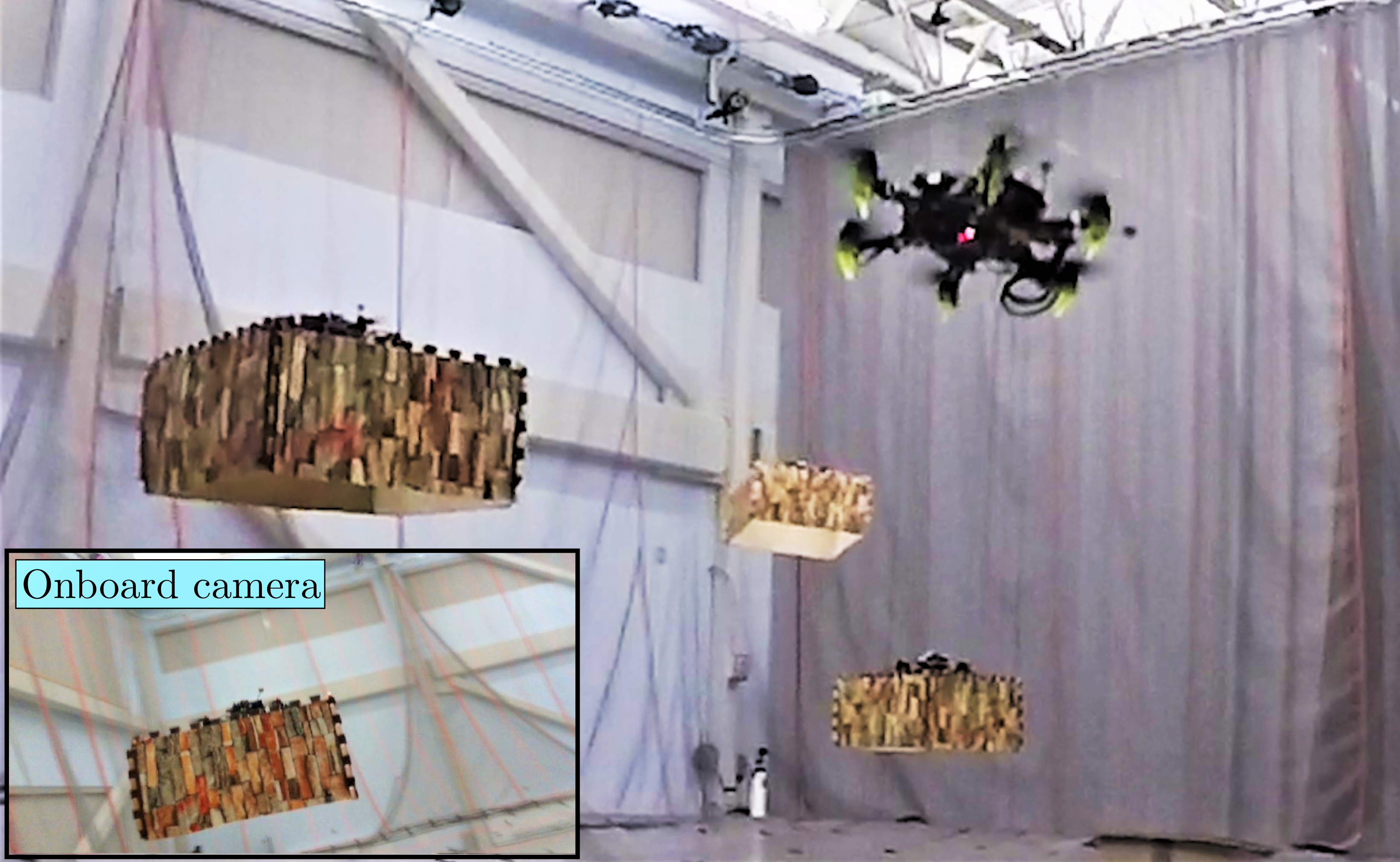}
		\par\end{centering}
	\caption{UAV planning perception-aware trajectories in a dynamic unknown environment, with relative velocities of up to $6.3$ m/s.
		All the computation runs onboard, and the UAV does not have any prior
		knowledge of the trajectories or specific shape/size of the dynamic
		obstacles. \label{fig:cover_photo_improved}}
\end{figure}

When a UAV equipped with a limited FOV sensor is flying in an
unknown environment \addrevision{(e.g.,~Fig~\ref{fig:cover_photo_improved})}, it is crucial to plan both the position and orientation of the UAV to maximize the detection and the tracking accuracy of the unknown obstacles while at the same time doing obstacle avoidance. This perception-aware (PA) component is especially important when flying in dynamic environments, because a consistent detection of the moving obstacles is necessary to obtain a good estimate of their locations and  prediction of their future trajectories.

Perception-awareness for UAVs has been studied thoroughly in the literature, and, as shown in Table~\ref{tab:classification}, the related work could be classified according to the formulation used and the goal itself. From the point of view of the \fcolorbox{black}{color_formulation!50}{\textbf{formulation}} used, there are approaches that are \textbf{not PA}~\circleform{1}, which typically plan the translation and then have either a constant yaw or a yaw such that the FOV of the camera points in the direction of travel~(e.g., see~\citenoPA{}). For instance,~\cite{falanga2020dynamic} used potential fields to avoid dynamic obstacles, but without taking into account perception-awareness, which can degrade the detection and prediction of the trajectories of the obstacles.

Other approaches are \textbf{PA by including additional hardware}~\circleform{2}: For example, by gimbal-mounting the camera, some of its degrees of 
freedom can be controlled independently of the rotation of the UAV \cite{nageli2017real,bonatti2018autonomous,chen2020bio}. Another option is to mount omni-directional sensors~\cite{ding2019efficient}. However, these approaches usually require additional hardware and mechanical complexity, which is typically undesirable on small UAVs. 

\textbf{PA planning}~\circleform{3} has received increased attention over the last few years due to its inherent ability to leverage the trajectory planned to maximize the PA objective. The related works could be subclassified according to whether or not the translation and yaw of the UAV are jointly optimized. On one hand, there are approaches that \uline{decouple} translation and yaw by optimizing them separately~\citePADec{}. For instance, in~\cite{zhou2020raptor}, a yaw trajectory is obtained for a fixed translational path to gain information about unknown static obstacles. For features or landmarks whose locations are known a priori, \cite{spasojevic2020perception} optimizes the time parametrization on a fixed spatial and yaw path to maximize their visibility. In~\cite{murali2019perception}, translation is optimized first, and then yaw is optimized to guarantee the co-visibility of the features. While this decoupling of translation and yaw has computational advantages, it can lead to conservative results, since the translational trajectory (and consequently two degrees of freedom of the rotation as well) is fixed in the yaw optimization. Other works assume a downward-facing camera, and hence only translation (not yaw) is planned to keep a specific target in the FOV of the camera~\cite{thomas2017autonomous}.

Another approach taken is to \uline{jointly optimize} translation and yaw, which enables the planner to fully exploit \emph{both} the
position trajectory and the yaw angle~\citePAJoint{}. This joint optimization leads to less conservative results than the approaches that decouple translation and yaw, but it typically comes at the expense of much higher computation times, especially when done in combination with dynamic obstacle avoidance constraints. For example,~\cite{watterson2020trajectory}
proposed an on-manifold trajectory optimization approach that couples
together translation with the full rotation, but the computation times
required (up to $30$~s) are not real time. Ref.~\cite{falanga2018pampc}
successfully presented a real-time MPC formulation that keeps the centroid of the VIO features in the center of the image while
minimizing its projected velocity. However, this formulation
does not include collision avoidance of static (or dynamic) obstacles, which greatly simplifies the
complexity of the optimization problem. In~\cite{penin2018vision}, translation and yaw are optimized jointly, but only static obstacle avoidance is performed. 
The technical gap then is how to jointly optimize the full pose of the UAV, satisfy its \addaccess{underactuated dynamics}, and guarantee safety in dynamic environments while maintaining real-time computational tractability.

The \addaccess{underactuated dynamics} of the UAV (caused by the \addaccess{total thrust of the UAV being fixed in the body frame})
makes this joint optimization especially hard, since a given spatio-temporal path fixes two degrees of freedom of the rotation, leaving only one extra degree of freedom in the rotation\footnote{Usually referred to as \emph{yaw}, \emph{heading}, or simply $\psi$}. A typical way to impose this constraint is via the dynamic equations of the UAV. However, this comes at the expense of having differential equations as constraints in the optimization. 

\newcommand{\diffFlatnessMap}{$(\mathbf{a}\in\mathbb{R}^3\setminus \arraycolsep=1.2pt\left[0 \; 0 \; -g\right]^{T},\psi\in S^1)\rightarrow\mathbf{R}_b^w \in \text{SO}(3)$}%

An alternative is to leverage the differential flatness of the UAVs~\cite{mellinger2011minimum} and make use of the map  \diffFlatnessMap{} that maps  $\psi$ and the acceleration $\mathbf{a}$ to the rotation of the body. However, and due to the \emph{hedgehog theorem} in $S^2$~\cite{bendixson1901courbes,brouwer1911abbildung}, there is no single continuous function that defines this map for all possible accelerations $\mathbf{a}$. For the most common definitions of this map, the singularity appears for each $\psi$ at two antipodal points in the unit sphere of possible normalized relative accelerations, which means that there is at least one singularity with a great-circle distance $\le 90^\circ$ with respect to the hovering condition. This closeness between the hovering condition and the singularity can limit the set of possible accelerations in aggressive flights, 
since an optimal solution that passes through or close to this singularity can provoke numerical instabilities and/or lead to artificial large changes in orientation.
Recently, the Hopf map was leveraged in~\cite{watterson2020control} to place the singularity in the inverted (``upside-down'') configuration, which is independent of $\psi$ and has the farthest possible angle away from the hovering condition. \addaccess{Although flying highly aggressive trajectories is not the main goal of this work, we decide to use the Hopf map (as opposed to the commonly-used maps presented in~\cite{mellinger2011minimum, faessler2017differential}) since it automatically maximizes the distance to the singularity by simply changing the definition of the map. 
In~\cite{watterson2020control}, however,} the Hopf fibration was only used in the controller to track predefined trajectories. It was also leveraged in~\cite{watterson2018geometric} to find the set of charts for a previously-optimized position trajectory, which are then used for the controller and to obtain the $\psi$ trajectory. 
In this work, we propose instead to embed the Hopf fibration in the joint (translation \emph{and} yaw) coupled planning optimization as a way to directly obtain trajectories in $\text{SE}(3)$ that, by construction, satisfy the \addaccess{underactuated dynamics} of the UAV.

From the point of view of the \fcolorbox{black}{color_goal!50}{\textbf{goal}} of the perception awareness, most of the related works focus on \textbf{reducing the state estimation uncertainty}~\circlegoal{1}, usually by keeping specific features/landmarks in the FOV, and/or choosing high-textured areas~\citeReduceStateEst{}. These features are typically static in the world frame. Some of these approaches also leverage the Observability Gramian \cite{preiss2018simultaneous,frey2019towards,salaris2019online}, especially when trying to ease the estimation of an unknown parameter of the dynamical system.

Further relevant work addresses the problem of having a UAV \textbf{record or chase a target}~\circlegoal{2} \citeRecordTarget{}. For example, \cite{thomas2017autonomous} focused on tracking a moving target with a downward-facing camera, while~\cite{penin2018vision} proposed a way to follow a moving target while avoiding other static obstacles in the environment. Most of these works focus therefore on chasing a static or dynamic target, not on avoiding it.

Our work differs from these two previous approaches because it proposes the use of PA planning to enhance the \textbf{avoidance of dynamic obstacles}~\circlegoal{3}. Compared to~\circlegoal{1} or~\circlegoal{2}, PA planning to avoid unknown dynamic obstacles comes with many additional challenges, such as the coupling of \emph{both} the ego-motion and the motion of the obstacle in the visibility cost and blur of the image, the inclusion of dynamic obstacle avoidance constraints in the optimization, the need to predict the future trajectories of the obstacles, and the consideration of the uncertainty of these predicted trajectories, just to name a few. 

\newcommand{\NextTraj}{\tikz[baseline=0.0ex]\draw [red,thick, dash pattern=on 3pt off 1pt] (0,0.08) -- (0.5,0.08);}
\newcommand{\CurrTraj}{\tikz[baseline=0.0ex]\draw [red,thick] (0,0.08) -- (0.5,0.08);}
\begin{table}
	\begin{centering}
		\caption{Notation used in this paper.\label{tab:Notation}}
		\vspace{-0.3cm}
		\par\end{centering}
	\noindent\resizebox{\columnwidth}{!}{%
		\begin{centering}
			\begin{tabular}{|>{\centering}m{0.31\columnwidth}|>{\raggedright}m{1.09\columnwidth}|}
				\hline 
				\textbf{Symbol} & \textbf{\qquad \qquad \qquad \qquad Meaning}\tabularnewline
				\hline 
				\hline 
				$\text{abs}\left(\boldsymbol{a}\right)$, $\boldsymbol{a}\le\boldsymbol{b}$&Element-wise absolute value, element-wise inequality. \tabularnewline
				\hline 
				$\left\Vert \cdot\right\Vert $&Euclidean norm. \tabularnewline
				\hline 
				$c_{\alpha}$, $s_{\alpha}$&$\text{cos(\ensuremath{\alpha})}$, $\text{sin}(\alpha)$ \tabularnewline
				\hline 
				$\circ$&Quaternion multiplication. \tabularnewline
				\hline 
				$g$&$g\approx9.81$~\accelunits{}\tabularnewline
				\hline 
				$\sigma(\cdot)$&Sigmoid function~\cite{sigmoid2020}. \tabularnewline
				\hline 
				$\boldsymbol{e}_{z}$, $\boldsymbol{1}$&$\arraycolsep=1.4pt\boldsymbol{e}_{z}:=\left[\begin{array}{ccc}0 & 0 & 1\end{array}\right]^{T}$, $\arraycolsep=1.4pt\boldsymbol{1}:=\left[\begin{array}{ccc}1 & 1 & 1\end{array}\right]^{T}$ \tabularnewline
				\hline 
				FOV, AABB&Field of View, Axis-Aligned Bounding Box. \tabularnewline
				\hline
				$\text{SO}(n)$, $\text{SE}(n)$&Special orthogonal group, Special Euclidean group. \tabularnewline
				\hline  
				$S^{n}$&$n$-sphere. \tabularnewline
				\hline 
				$\text{wrap}_{-\pi}^{\pi}(\cdot)$&Wrapping of an angle in $[-\pi,\pi)$ \tabularnewline
				\hline 
				$N(\cdot)$&Normal distribution. \tabularnewline
				\hline 
				$\text{norminv}(\cdot)$&Inverse of the standard normal cumulative distribution function \cite{norminv21}. \tabularnewline
				\hline 
				$\text{diag}(\cdot)$&Diagonal matrix.\tabularnewline
				\hline 
				$\mathcal{S}_{p,m}^{d}$  & Set of clamped uniform splines with dimension $d$, degree $p$, and
				$m+1$ knots.\tabularnewline
				\hline 
				$n$ ($n_{\mathbf{p}}$ and $n_{\psi}$ ) & $n:=m-p-1$
				
				$n+1$ is the number of control points of the spline.\tabularnewline
				\hline 
				$\mathbf{p},\mathbf{v},\mathbf{a},\mathbf{j}$ & Position, Velocity, Acceleration\addrevision{,} and Jerk \addrevision{of the UAV}, $\in\mathbb{R}^{3}$. All
				of them are of the body w.r.t. the world frame, and expressed in the
				world frame.\tabularnewline
				\hline 
				$\boldsymbol{\xi}$ & Relative acceleration, expressed in the world frame: $\arraycolsep=2.1pt\boldsymbol{\xi}:=\left[\begin{array}{ccc}
					\mathbf{a}_{x} & \mathbf{a}_{y} & \mathbf{a}_{z}+g\end{array}\right]^{T}$. We will assume $\boldsymbol{\xi}\neq\boldsymbol{0}$.\tabularnewline
				\hline 
				$\psi$, $\dot{\psi}$ & Angle (and its derivative) such that $\arraycolsep=1.4pt\boldsymbol{q}_{b}^{w}=\boldsymbol{q}_{\boldsymbol{\xi}}\circ\left[\begin{array}{cccc}
					c_{\psi/2} & 0 & 0 & s_{\psi/2}\end{array}\right]^{T}$ (see \addaccess{section~\ref{sec:Coupling-rotation-and}}).\tabularnewline
				\hline 
				$\mathbf{x}$ & State vector: $\arraycolsep=1.4pt\mathbf{x}:=\left[\begin{array}{ccccc}
					\mathbf{p}^{T} & \mathbf{v}^{T} & \mathbf{a}^{T} & \psi & \dot{\psi}\end{array}\right]^{T}\in\mathbb{R}^{11}$.\tabularnewline
				\hline 
				$\boldsymbol{p}^{a}$ & Point expressed in the frame $a$. For the definitions of this table
				that include the sentence \emph{``expressed in the world frame''},
				the notation of the frame is omitted.\tabularnewline
				\hline 
				$\tilde{\boldsymbol{p}}$, $\boldsymbol{\bar{p}}$ & $\arraycolsep=1.4pt\tilde{\boldsymbol{p}}:=\left[\begin{array}{cc}
					\boldsymbol{p}^{T} & 1\end{array}\right]^{T}$, $\boldsymbol{\bar{p}}:=\frac{\boldsymbol{p}}{\left\Vert \boldsymbol{p}\right\Vert }$\tabularnewline
				\hline 
				$\arraycolsep=1.4pt\boldsymbol{T}_{b}^{a}=\left[\begin{array}{cc}
					\boldsymbol{R}_{b}^{a} & \boldsymbol{t}_{b}^{a}\\
					\boldsymbol{0}^{T} & 1
				\end{array}\right]$ & Transformation matrix: $\tilde{\boldsymbol{p}}^{a}=\boldsymbol{T}_{b}^{a}\tilde{\boldsymbol{p}}^{b}$.
				Analogous definition for the quaternion $\boldsymbol{q}_{b}^{a}$.\tabularnewline
				\hline 
				$\text{rot\ensuremath{\left(\boldsymbol{q}\right)}}$ & Rotation matrix associated with the quaternion $\boldsymbol{q}$.\tabularnewline
				\hline 
				$J$ & Set of indexes of all the intervals $J=\{0,1,...,m-2p-1\}$.\tabularnewline
				\hline 
				$j$ & Index of the interval of the trajectory, $j\in J$.\tabularnewline
				\hline 
				$I$ & Set of indexes of the tracked obstacles.\tabularnewline
				\hline 
				$i$ & Index of the obstacle, $i\in I$.\tabularnewline
				\hline 
				$i^{*}$ & Index of the obstacle used in the PA term of the cost function.\tabularnewline
				\hline 
				\addrevision{$\left(\mathbf{p}_{i}\right)^{a}(t)$}&\addrevision{Mean of the predicted position of obstacle $i$, expressed in frame~$a$}\tabularnewline
				\hline 
				$\left(\mathbf{p}_{i}\right)^{w}(t)$, $\mathbf{\boldsymbol{\sigma}}_{i}(t)$ & The predicted trajectory of the obstacle $i$, in the world frame,
				is {\normalsize{}$\sim N\left(\left(\mathbf{p}_{i}\right)^{w}(t),\left(\text{diag}\left(\mathbf{\boldsymbol{\sigma}}_{i}(t)\right)\right)^{2}\right)$}.\tabularnewline
				\hline 
				$f$  & Focal length of the camera in meters.\tabularnewline
				\hline 
				$\theta$ & Opening angle of the cone that approximates the FOV.\tabularnewline
				\hline \vspace{0.05cm}
				$\arraycolsep=1.4pt\left[\begin{array}{cccc}
					q_{w} & q_{x} & q_{y} & q_{z}\end{array}\right]^{T}$ & Components of a unit quaternion.\tabularnewline
				\hline 
				$\boldsymbol{1}_{c}$ & $1$ if $c$ is true, 0 otherwise.\tabularnewline
				\hline 
				$\text{inFOV}\left(\boldsymbol{T}_{c}^{w},\left(\mathbf{p}_{i}\right)^{w}\right)$ & $\boldsymbol{1}_{\ensuremath{\left(\mathbf{p}_{i}\right)^{w}\in}\text{FOV}}\approx\boldsymbol{1}_{\left(\left(\mathbf{p}_{i}\right)^{c}\right)_{z}/\left\Vert \left(\mathbf{p}_{i}\right)^{c}\right\Vert \ge\ensuremath{c_{\theta/2}}}\approx\sigma\left(\gamma\left(-\ensuremath{c_{\theta/2}}+\left(\left(\mathbf{p}_{i}\right)^{c}\right)_{z}/\left\Vert \left(\mathbf{p}_{i}\right)^{c}\right\Vert \right)\right)$. $\gamma$ is a positive parameter.\tabularnewline
				\hline 
				$L_{\mathbf{p}}$, $L_{\psi}$ & $L_{\mathbf{p}}:=\{0,1,...,n_{\mathbf{p}}\}$, $L_{\psi}:=\{0,1,...,n_{\psi}\}$.\tabularnewline
				\hline 
				$l$ & Index of the control point. $l\in L_{\mathbf{p}}$ for $\mathbf{p}(t)$,
				$l\in L_{\mathbf{p}}\backslash\{n_{\mathbf{p}}\}$ for $\mathbf{v}(t)$, $l\in L_{\mathbf{p}}\backslash\{n_{\mathbf{p}}-1,n_{\mathbf{p}}\}$
				for $\mathbf{a}(t)$, $l\in L_{\psi}$ for $\psi$ and $l\in L_{\psi}\backslash\{n_{\psi}\}$
				for $\dot{\psi}$\tabularnewline
				\hline 
				$\boldsymbol{q}_{l},\boldsymbol{v}_{l},\boldsymbol{a}_{l},\psi_{l},\Psi_{l}$ & Position, velocity, and acceleration control points, $\left(\in\mathbb{R}^{3}\right)$,
				$\psi$ and $\dot{\psi}$ control points $\left(\in\mathbb{R}\right)$.\tabularnewline
				\hline 
				$\mathcal{Q}_{j}^{\text{MV}}$ & Set of position control points of the interval $j$ using the MINVO
				basis. Analogous definition for the velocity control points $\mathcal{V}_{j}^{\text{MV}}$.\tabularnewline
				\hline 
				$\delta$ & $\in[0,1]$, percentile of the standard normal distribution (see next
				row).\tabularnewline
				\hline 
				$\mathcal{C}_{ij}^{\text{MV}}$ & Set of vertexes of the convex hull of the set obtained by inflating
				{\normalsize{}$\left(\mathcal{Q}_{j}^{\text{MV}}\right)_{\text{obs i}}$}
				with {\normalsize{}$\text{norminv}(\delta)\cdot\mathbf{\boldsymbol{\sigma}}_{i}\left(t_{\text{end \ensuremath{j}}}\right)$},
				half of the sides of the AABB of the obstacle $i$ and half of the
				sides of the AABB of the agent. \tabularnewline
				\hline 
				$\boldsymbol{\pi}_{ij}$ ($\boldsymbol{n}_{ij}$, $\;d_{ij}$) & Plane $\boldsymbol{n}_{ij}^{T}\boldsymbol{x}+d_{ij}=0$ that separates
				$\left(\mathcal{Q}_{j}^{\text{MV}}\right)_{\text{agent}}$ from $\mathcal{C}_{ij}^{\text{MV}}$.\tabularnewline
				\hline 
				$\boldsymbol{h}(\cdot)$, $\boldsymbol{st}(\cdot)$  & Hopf fibration, stereographic projection.\tabularnewline
				\hline 
				\multicolumn{2}{|l|}{Snapshot at $t=t_{1}$ (current time):}\tabularnewline
				\multicolumn{2}{|c|}{\includegraphics[width=0.6\columnwidth]{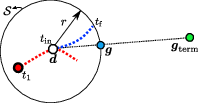}}\tabularnewline
				\multicolumn{2}{|l|}{$\boldsymbol{g}_{\text{term}}$ (\tikzcircle[black,fill=green]{15pt})
					is the terminal goal, and \tikzcircle[black,fill=red]{15pt} is the
					current position of the UAV.}\tabularnewline
				\multicolumn{2}{|l|}{\CurrTraj \ is the trajectory the UAV is currently executing.}\tabularnewline
				\multicolumn{2}{|l|}{\NextTraj \ is the trajectory the UAV is currently optimizing, $t\in\left[t_{\text{in}}, t_{\text{f}}\right]$} \tabularnewline
			\multicolumn{2}{|l|}{$\boldsymbol{d}$ (\tikzcircle[black,fill=white]{15pt}) is a point
				in \CurrTraj, used as the initial position of \NextTraj}\tabularnewline
			\multicolumn{2}{|l|}{$\mathcal{M}$ is a sphere of radius $r$ around $\boldsymbol{d}$.}\tabularnewline
			\multicolumn{2}{|l|}{$\boldsymbol{g}$ (\tikzcircle[black,fill=blue_light]{15pt}) is the
				projection of $\boldsymbol{g}_{\text{term}}$ (\tikzcircle[black,fill=green]{15pt}) onto the sphere $\mathcal{M}$.}\tabularnewline
			\multicolumn{2}{|l|}{\addrevision{$\boldsymbol{d}$, $\boldsymbol{g}$, and $\boldsymbol{g}_\text{term}$ are expressed in the world frame.}}\tabularnewline
			\hline 
		\end{tabular}
		\par\end{centering}
}
\end{table}

In summary, the proposed contributions of this work are as follows:
\begin{itemize}
\item Real-time PA planning formulation that jointly optimizes the translation and the full rotation to maximize the visibility of unknown dynamic obstacles, while simultaneously avoiding them. Compared to non-PA approaches and PA decoupled approaches, our proposed coupled solution leads to a presence of the obstacle in the FOV \addaccess{7.9} and \addaccess{1.5} times more frequent, respectively. The success rates achieved are \addaccess{on average $2.98$} times larger than other state-of-the-art approaches when flying in multi-obstacle dynamic environments. 

\item We show how the Hopf fibration can be embedded in the planning optimization to jointly optimize translation and yaw while implicitly imposing the \addaccess{underactuated dynamics} that couples acceleration and orientation. This avoids the need to explicitly impose the dynamics of the UAV as differential constraints\addaccess{, while automatically guaranteeing} the largest possible great-circle distance between the hovering condition and the differential flatness singularity. Dynamic obstacle avoidance constraints are imposed by leveraging the MINVO basis to reduce conservatism.

\item Extensive set of hardware experiments in unknown dynamic environments,
with everything (navigation, perception, planning, and control) executed onboard the UAV, and without any prior knowledge of the trajectories or specific shape/size of the obstacles. The UAV achieves velocities of up to 5.8~m/s and relative velocities (with respect to the obstacles) of up to~6.3~m/s. The replanning times achieved onboard are~$\approx 53$~ms. 
\item The code has also been released open source for the community.
\end{itemize}

This paper uses the notation shown in Table \ref{tab:Notation}.

\section{PANTHER}\label{sec:panther}

\begin{figure*}
	\sidesubfloat[]{\includegraphics[width=0.98\textwidth]{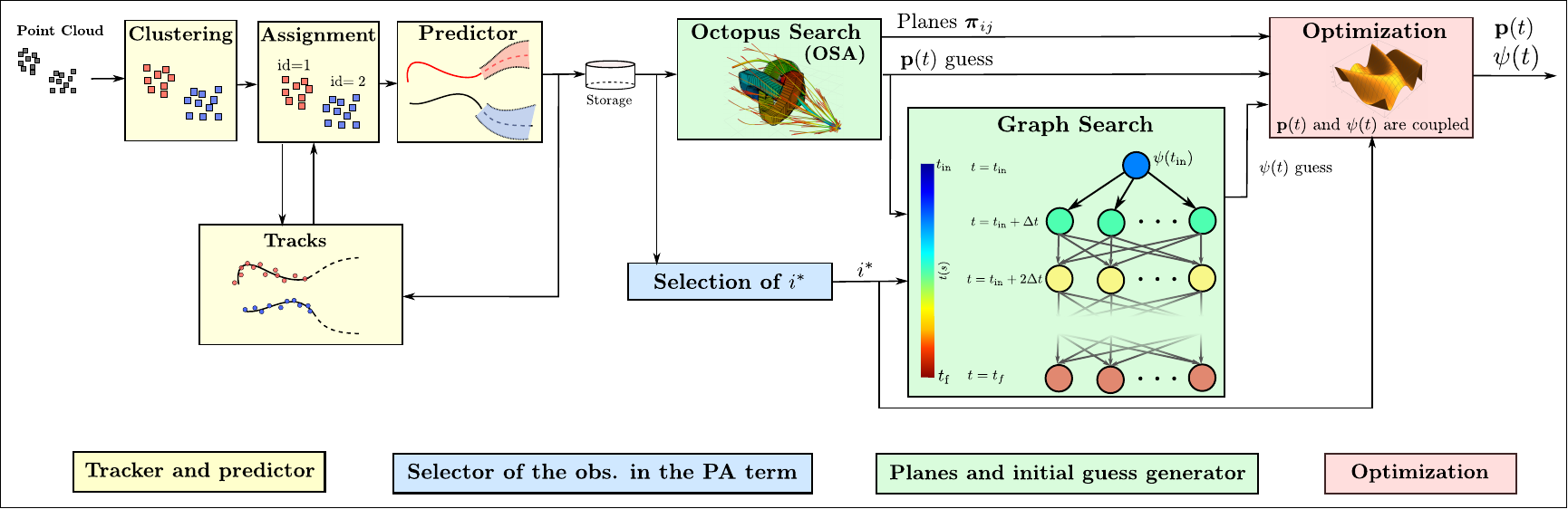}\label{fig:full_diagram}}\quad
	\vskip 0.5cm
	\sidesubfloat[]{\includegraphics[width=0.48\textwidth]{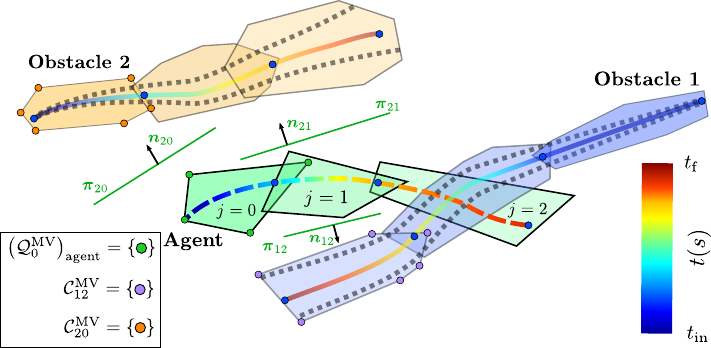}\label{fig:explanation_trajs}}%
	\sidesubfloat[]{\includegraphics[width=0.48\textwidth]{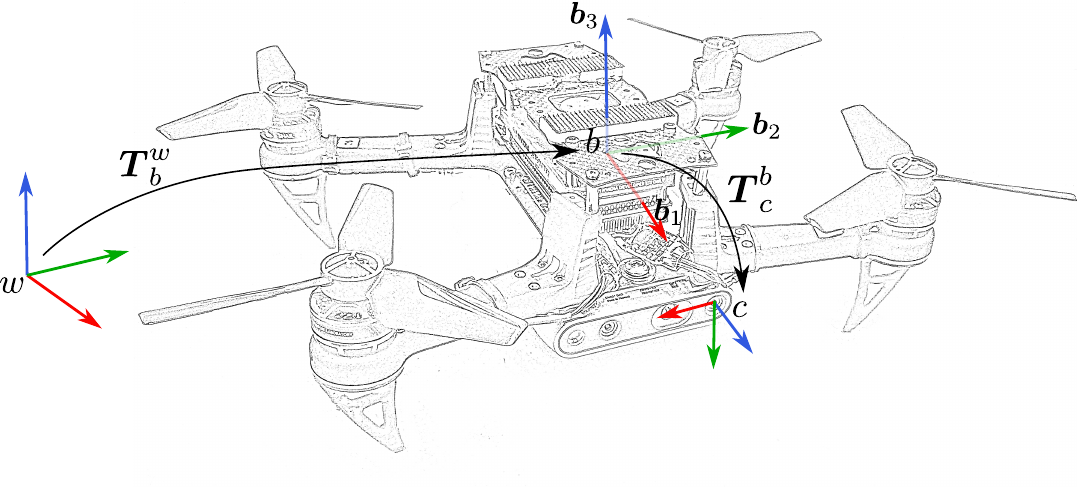}\label{fig:drone_jetson_coordinates}}%
	\vskip 0.5cm
	\sidesubfloat[]{\includegraphics[width=0.98\textwidth]{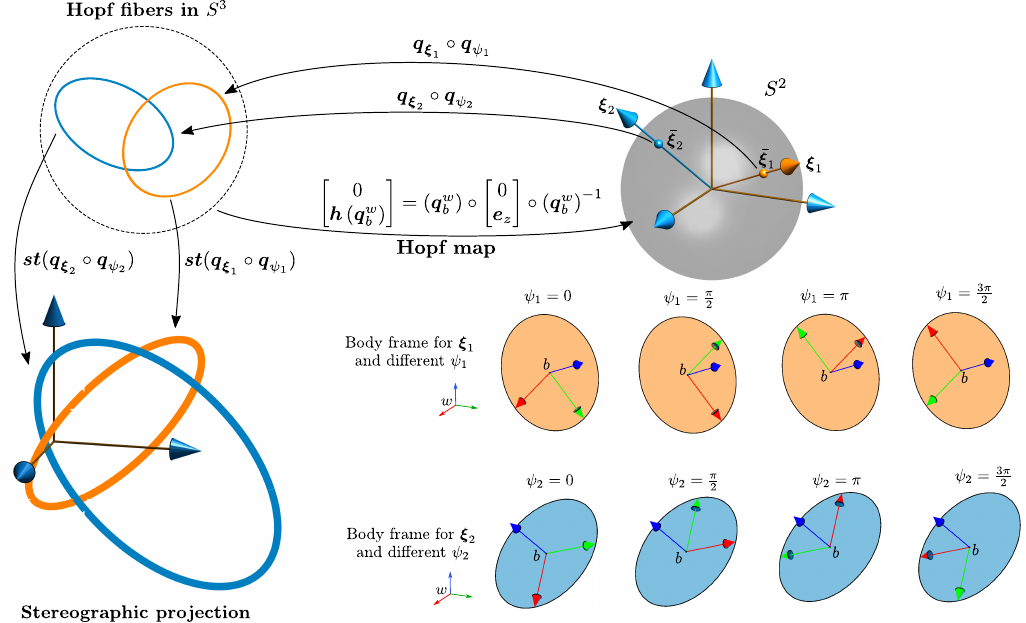}\label{fig:hopf}}%
	\caption{ (\textbf{A}) Different modules of PANTHER. (\textbf{B}) Predicted trajectories of the obstacles and convex representation of each segment of the trajectory of the agent and the obstacles.  (\textbf{C}) World, body, and camera frames. (\textbf{D}) Hopf fibration and its stereographic projection, partly inspired from \cite{lyons2003elementary}. Given a specific relative acceleration $\boldsymbol{\xi}$ (with $\bar{\boldsymbol{\xi}}\neq-\boldsymbol{e}_{z}$), the quaternion $\boldsymbol{q}_{b}^{w}=\boldsymbol{q}_{\boldsymbol{\xi}}\circ\boldsymbol{q}_{\psi}$
		is a fiber (specifically a circle) in $S^{3}$ parameterized
		by $\psi$. On the bottom right, the body frames for different values of $\psi_i$ for each $\boldsymbol{\xi}_i$ are shown.  }%
\end{figure*}

PANTHER comprises four modules: Tracker and predictor, selector of the obstacle in the PA term, planes and initial guess generator, and optimization (see Fig.~\ref{fig:full_diagram}). 
\addrevision{A summary of how all these modules work together is as follows: First the incoming point clouds of the onboard depth sensor are clustered and tracked using the Hungarian algorithm~\cite{kuhn1955hungarian} to obtain the trajectory, as a probability distribution, of each of the obstacles (section~\ref{sec:tracker_and_prediction}). The obstacle $i^*$ that the UAV is most likely to collide with is then selected to be included in the PA term of the cost function~(section~\ref{sec:several_obstacles}). Then, a kinodynamic search-based planner (Octopus Search Algorithm~\cite{tordesillas2020mader}) is run to find a initial guess of the translational  trajectory $\mathbf{p}(t)$ that avoids the probabilistic trajectories of the obstacles found before (section~\ref{subsec:init_guess_pos}). This translational guess and the obstacle $i^*$ selected are then used to run a graph search algorithm to find the $\psi(t)$ guess (section~\ref{subsec:init_guess_psi}). Finally, the $\mathbf{p}(t)$ and $\psi(t)$ guesses are used for the nonconvex optimization to obtain the optimized trajectory, that is sent to the controller of the UAV (section~\ref{sec:optimization}). In this framework, the coupling between rotation and acceleration is imposed implicitly using the Hopf fibration. All these} modules are described in detail in the following subsections.

\subsection{Tracking and Prediction}\label{sec:tracker_and_prediction}
We create a k-d tree representation of the point clouds coming from the onboard depth sensor, and perform Euclidean clustering to group the points that are more likely to belong to the same obstacle (see Fig.~\ref{fig:full_diagram})\addaccess{. For each cluster found, we compute the AABB (Axis-Aligned Bounding Box) centered on the centroid of that cluster}\footnote{\addaccess{Regardless of whether or not the obstacle is convex, this produces an outer convex approximation of the visible part of the obstacle.}}\addaccess{. Then, to} assign each cluster to a specific track, we minimize the total assignment cost using the Hungarian algorithm~\cite{kuhn1955hungarian}, where the cost is the pairwise distance between the centroid of each cluster and the prediction of the tracks at the time the point cloud was produced. If this distance is above a specific threshold (usually $\approx 1$--2 m), we create a new track for it. If a cluster is not assigned to any track (which can happen if there are more clusters than tracks), then a new track is created for it. Finally, given a sliding window history of all the observations associated with a track, we fit a polynomial for each coordinate $\{x,y,z\}$. \addrevision{To capture the stochasticity of the prediction problem, t}he predicted position at time $t$ is then approximated by a 3D Gaussian distribution (mean from the value of the fitted polynomial and a diagonal covariance matrix obtained from the prediction intervals~\cite[\addrevision{section}~5.7]{hyndman2018forecasting}).

\subsection{Selection of the obstacle in the PA term}\label{sec:several_obstacles}
When there are several predicted trajectories, and to maintain computational tractability, the agent needs to choose which one of them to include in the PA term of the cost function. It does so by choosing the most likely obstacle to collide with in the future, using a simple heuristic of the probability of collision based on Boole's inequality \cite{janson2018monte}:
\[
i^{*}=\underset{i\in I}{\text{argmax}}\sum_{u=0}^{U-1}P\left(\left\Vert \left(\mathbf{p}_{i}\right)^{\addrevision{w}}\left(t_{u}\right)-\boldsymbol{\kappa}\left(u\right)\right\Vert _{\infty}\le R\right)
\]
where $U$ is the number of samples taken, $t_u:=t_{\text{in}}+\frac{u}{U}(t_{f}-t_{\text{in}})$ and $\boldsymbol{\kappa}(u):=\boldsymbol{d}+\frac{u}{U}(\boldsymbol{g}_{\text{term}}-\boldsymbol{d})$
is a point in a straight line from $\boldsymbol{d}$ to $\boldsymbol{g}_{\text{term}}$.
Note that although only one obstacle is included in the PA 
objective function, all the predictions of the tracked 
obstacles are included in the collision avoidance constraints.

Additionally, and to address the trade-off between gathering information about the obstacle, and gathering information about the direction of travel, the UAV will include the obstacle~$i^{*}$ in the PA term if the angle between $\left(\boldsymbol{g}_\text{term}-\boldsymbol{d}\right)$ and $\left(\left(\boldsymbol{p}_{i^{*}}\right)^{\addrevision{w}}(t_\text{in})-\boldsymbol{d}\right)$ is smaller than a predefined angle $\alpha_0$ (typically $\approx 90^\circ$). Otherwise the UAV will try to align the FOV of the camera with the direction of travel.

\subsection{Planes and Initial guesses}

\subsubsection{Separability planes and initial guess for position}\label{subsec:init_guess_pos}

We use the Octopus Search Algorithm \addaccess{(OSA)~\cite{tordesillas2020mader}}, which is a search-based algorithm that operates directly on the control points of the position spline. It ensures collision-free constraints between the agent and the dynamic obstacles by finding the planes that separate the inflated MINVO polyhedral representation of each interval $j$ of the trajectory of the obstacle $i$ (denoted as $\mathcal{C}_{ij}^{\text{MV}}$) and the MINVO polyhedral representation of that interval $j$ of the trajectory of the agent, denoted as $\left (\mathcal{Q}_{j}^{\text{MV}}\right)_\text{agent}$ (see Fig.~\ref{fig:explanation_trajs}). The outputs of this algorithm are both the position control points and the planes $\boldsymbol{\pi}_{ij}$ (given by $\boldsymbol{n}_{ij}^{T}\boldsymbol{x}+d_{ij}=0$)~$\forall i,\forall j$. The position control points are then used as initial guess in the optimization, while the planes $\boldsymbol{\pi}_{ij}$ are held fixed in the optimization.
The reader is referred to our previous work~\cite{tordesillas2020mader} for a more in-depth explanation of the OSA.

\subsubsection{Initial guess for $\psi$}\label{subsec:init_guess_psi}

To obtain the initial guess for $\psi$, we uniformly sample the position
guess spline obtained through the OSA, and for
each of these position samples, we uniformly sample several values of $\psi\in[-\pi,\pi)$. Each one of these $\mathbf{p}$-$\psi$ samples will be a \textit{node}, and all the nodes associated with the same position sample, but with different $\psi$, will constitute a \emph{layer} (see Fig.~\ref{fig:full_diagram}). Then, we create a graph connecting with directed edges all the nodes of one layer 
to the nodes of the next layer \cite{zhou2020raptor}. Each node has therefore a time, position,
acceleration, and yaw associated with it, and all the nodes of the same layer have the same time, position, and acceleration. The cost of the edge between two nodes $n_{1}$ and
$n_{2}$ of the graph is then given by 
\begin{multline*}
	c_{\psi}\left(\text{wrap}_{-\pi}^{\pi}\left(\psi_{n_{2}}-\psi_{n_{1}}\right)\right)^{2} +c_{\Psi_{\text{max}}}\cdot\boldsymbol{1}_{\left|\frac{\text{wrap}_{-\pi}^{\pi}\left(\psi_{n_{2}}-\psi_{n_{1}}\right)}{t_{n_2}-t_{n_1}}\right|>\Psi_{\text{max}}} \\
	+c_{\text{FOV}}\left(1-\text{inFOV}\left(\left(\boldsymbol{T}_{n_{2}}\left(t_{n_{2}}\right)\right)_{c}^{w},\left(\boldsymbol{p}_{i^{*}}\left(t_{n_{2}}\right)\right)^{w}\right)\right)
\end{multline*}

Here, $c_{\psi}$, $c_{\Psi_{\text{max}}}$, and $c_\text{FOV}$ are nonnegative weights, while $\psi_{n_{u}}$, $t_{n_{u}}$, and $\left(\boldsymbol{T}_{n_{u}}\left(t_{n_{u}}\right)\right)_{c}^{w}$ are the angle $\psi$, the time, and the transformation matrix
associated with node $n_{u}$. Note that the edge cost is guaranteed to be nonnegative at all times. 
The transformation matrix can be directly obtained from the position, acceleration, and yaw of the node. The first term in the cost penalizes the distance between two $\psi$ angles, the second term penalizes edges that do not satisfy
the limit $\Psi_{\text{max}}$, and the last one rewards the visibility of the obstacle. 
\addrevision{The units of the weights above are such that the corresponding term is dimensionless (see section~\ref{sec:results_and_discussion}). To choose these weight values, we first set $c_{\Psi_{\text{max}}}$ to a large value to guarantee the ${\Psi_{\text{max}}}$ constraint. Then, $c_\psi$ and $c_\text{FOV}$ are selected as a trade-off between smoothness and inclusion of the obstacle $i^*$ in the FOV of the UAV.} 
The root node of the graph corresponds to the state  %
$\boldsymbol{d}$ (see last row of Table~\ref{tab:Notation}). 
We solve the search problem using Dijkstra's algorithm~\cite{dijkstra1959note},
with early termination when the search reaches a node of
the last layer. Letting $\Lambda$ denote the indexes of the nodes of the path found, we shift the angles $\psi_{n_\lambda} \; \forall \lambda \in \Lambda$ (by adding or subtracting $2\pi r$, $r\in\mathbb{{Z}}$) such
that the absolute difference between two consecutive angles is $\le\pi$. Using $\hat\psi_{n_\lambda}$ to denote these shifted angles, a spline is fitted to these angles by solving 
the following constrained least square problem: 

\begin{align}\label{eq:fit_yaw_samples}
	\begin{split}
		&\underset{\psi(t)\in\mathcal{S}_{2,m}^{1}}{\boldsymbol{\min}}\sum_{\lambda\in \Lambda}\left\Vert \psi\left(t_{n_\lambda}\right)-\hat\psi_{n_\lambda}\right\Vert _{2}^{2}  \\
		&\text{s.t.} \quad \psi(t_{\text{in}})=\psi_{\text{in}}, \quad\dot{\psi}(t_{\text{in}})=\dot{\psi}_{\text{in}}, \quad\dot{\psi}(t_{\text{f}})=0
	\end{split}
\end{align}

\addrevision{Note that, as this problem is a quadratic program with linear equality constraints, its solution can be easily found by simply solving the linear Karush-Kuhn-Tucker (KKT) conditions associated with it \cite{kuhn1951nonlinear, karush1939minima}\footnote{\addrevision{For a detailed explanation of the derivation of the resulting linear system of equations, see, e.g., \cite[Example 5.1]{boyd2004convex}.}}. }
The control points of this fitted spline are then used as the initial guess for $\psi(t)$ in the optimization. 

\subsection{Optimization}\label{sec:optimization}

\begin{table*}[t]
	\begin{centering}
		\caption{Some commonly-used definitions for the differential flatness map \diffFlatnessMap{}. The colormap represents the great-circle distance to the closest singularity (yellow is closer), $(\cdot)_n$ denotes the normalization of a vector, and $\arraycolsep=2.1pt\bar{\boldsymbol{\xi}}:=(\left[
			\mathbf{a}_{x} \;\; \mathbf{a}_{y} \;\; \mathbf{a}_{z}+g\right]^{T})_n$
			is the normalized relative acceleration, expressed in the world frame. See also \cite{morrell2018differential,watterson2018geometric,spasojevic2020perception,spica2013open,allen2019real} for more possible definitions, which are usually rotations of the first two definitions of this table. \label{tab:singularities}}
		\noindent\resizebox{\textwidth}{!}{%
			\begin{tabular}{|>{\centering}m{0.09\textwidth}|>{\centering}m{0.28\textwidth}|>{\centering}m{0.28\textwidth}|>{\centering}m{0.35\textwidth}|}
				\hline
				& \textbf{Definition 1} & \textbf{Definition 2} & \textbf{Definition 3 (Hopf \addrevision{f}ibration)}\tabularnewline 
				\hline
				\hline
				\vspace{1.0cm}\textbf{Map} &
				{$\arraycolsep=1.1pt \! \begin{aligned}
						&\boldsymbol{b}_{1}      = \boldsymbol{b}_{2}\times\boldsymbol{b}_{3}                                          \\
						&	\boldsymbol{b}_{2}      = \left(\boldsymbol{b}_{3}\times\left[\begin{array}{ccc}
							c_{\psi} & s_{\psi} & 0\end{array}\right]^{T}\right)_{n} \\
						&\boldsymbol{b}_{3}      =\bar{\boldsymbol{\xi}}                                       \\
						&\boldsymbol{R}_{b}^{w}  =\left[\begin{array}{ccc}
							\boldsymbol{b}_{1} & \boldsymbol{b}_{2} & \boldsymbol{b}_{3}\end{array}\right]
					\end{aligned}$}
				&
				{$\arraycolsep=1.1pt \!\begin{aligned}
						&\boldsymbol{b}_{1}      =\left(\left[\begin{array}{ccc}
							-s_{\psi} & c_{\psi} & 0\end{array}\right]^{T}\times\boldsymbol{b}_{3}\right)_{n} \\
						&\boldsymbol{b}_{2}  =\boldsymbol{b}_{3}\times\boldsymbol{b}_{1}                         \\
						&\boldsymbol{b}_{3}      =\bar{\boldsymbol{\xi}}                                     \\
						&\boldsymbol{R}_{b}^{w}  =\left[\begin{array}{ccc}
							\boldsymbol{b}_{1} & \boldsymbol{b}_{2} & \boldsymbol{b}_{3}\end{array}\right]
					\end{aligned}$}
				&
				\vspace{0.1cm}
				{$\arraycolsep=1.1pt \!\begin{aligned}
						\arraycolsep=1.1pt  \boldsymbol{q}_{b}^{w}&=\underbrace{\frac{1}{\sqrt{2(1+\bar{\boldsymbol{\xi}}_{z})}}\left[\begin{array}{c}
								1+\bar{\boldsymbol{\xi}}_{z}\\
								-\bar{\boldsymbol{\xi}}_{y}\\
								\bar{\boldsymbol{\xi}}_{x}\\
								0
							\end{array}\right]}_{:=\boldsymbol{q}_{\boldsymbol{\xi}}}\circ\underbrace{\left[\begin{array}{c}
								c_{\psi/2}\\
								0\\
								0\\
								s_{\psi/2}
							\end{array}\right]}_{:=\boldsymbol{q}_{\psi}} \\
						\boldsymbol{R}_{b}^{w}                   & =\text{rot}\left(\boldsymbol{q}_{b}^{w}\right)
					\end{aligned}$}
				\tabularnewline
				\hline
				\textbf{Singularity} &	\vspace{0.1cm}			$\arraycolsep=1.4pt\bar{\boldsymbol{\xi}}\parallel\left[\begin{array}{ccc}
					c_{\psi} & s_{\psi} & 0\end{array}\right]^{T}$. When $\psi=0$:\hspace{1cm}
				\includegraphics[width=2cm]{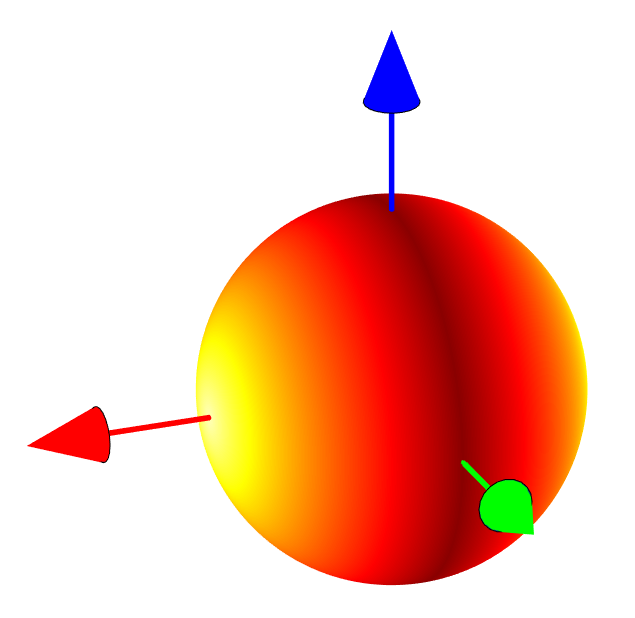} & \vspace{0.1cm}	
				$\arraycolsep=1.4pt\bar{\boldsymbol{\xi}}\parallel\left[\begin{array}{ccc}
					-s_{\psi} & c_{\psi} & 0\end{array}\right]^{T}$. When $\psi=0$:
				
				\includegraphics[width=2cm]{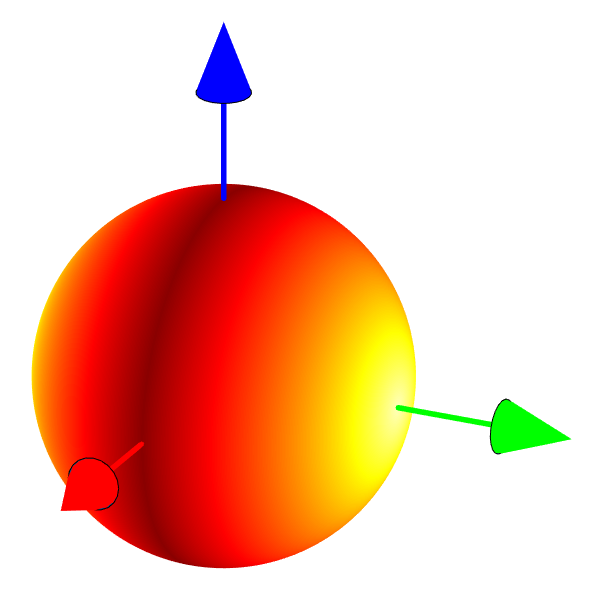} & \vspace{0.1cm}	 $\arraycolsep=1.4pt\bar{\boldsymbol{\xi}}=\left[\begin{array}{ccc}
					0 & 0 & -1\end{array}\right]^{T}$
				
				$\quad$
				\includegraphics[width=2cm]{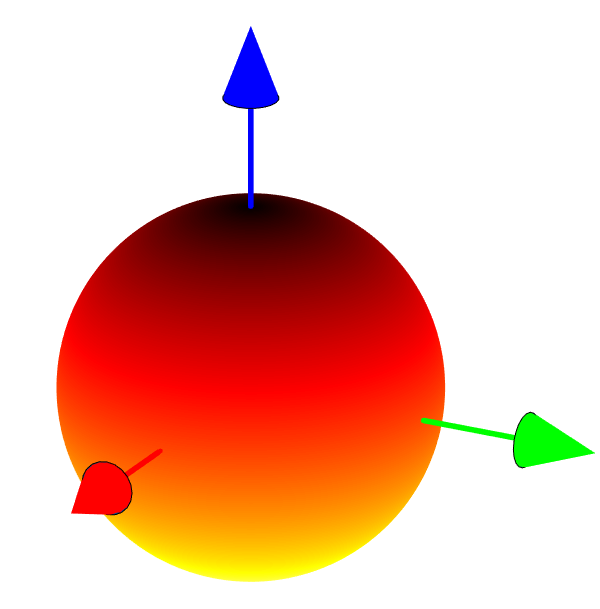}\tabularnewline
				\hline
				\textbf{Notes}			
				&  \vspace{0.2cm}\begin{itemize}[leftmargin=0.2cm]
					\item Singularity=$f(\mathbf{a},\psi)$
					\item For a given $\psi$, singularity appears for two $\bar{\boldsymbol{\xi}}$%
					\item UAV is differentially flat, with flat outputs $\{\mathbf{p},\psi\}$~\cite{mellinger2011minimum}
				\end{itemize} 
				&    \vspace{0.2cm}\begin{itemize}[leftmargin=0.2cm]
					\item Singularity=$f(\mathbf{a},\psi)$
					\item For a given $\psi$, singularity appears for two $\bar{\boldsymbol{\xi}}$%
					\item UAV is differentially flat, with flat outputs $\{\mathbf{p},\psi\}$~\cite{faessler2017differential}
				\end{itemize} &     \begin{itemize}[leftmargin=0.2cm]
					\item Singularity=$f(\mathbf{a})$
					\item Singularity appears for one $\bar{\boldsymbol{\xi}}$%
					\item UAV is differentially flat, with flat outputs $\{\mathbf{p},\psi\}$~\cite{watterson2020control}
				\end{itemize}\tabularnewline
				\hline
		\end{tabular}}
		\par\end{centering}
\end{table*}

\subsubsection{Coupling rotation and acceleration with the Hopf fibration \label{sec:Coupling-rotation-and}}
In a \addaccess{standard multirotor-}UAV, the perpendicularity of the \addaccess{total thrust} with respect to the plane spanned by  $\boldsymbol{b}_1$ and  $\boldsymbol{b}_2$ (see the coordinate frames shown in Fig.~\ref{fig:drone_jetson_coordinates}) \addaccess{makes the UAV underactuated by imposing the following constraint}~\cite{watterson2020control}:
\begin{equation} \label{eq:accel_rotation_constraint}
	\text{rot}\left(\boldsymbol{q}_{b}^{w}\right)\boldsymbol{e}_z=\bar{\boldsymbol{\xi}}
\end{equation}
where $\bar{\boldsymbol{\xi}}$ is the normalized relative acceleration expressed in the world frame (see Table~\ref{tab:Notation}). In a planning optimization problem where rotation and translation are jointly \addaccess{optimized,~\eqref{eq:accel_rotation_constraint} needs} to be satisfied at all times.  %
A very common way to \addaccess{guarantee~\eqref{eq:accel_rotation_constraint}} is via direct imposition of the dynamic equations of the UAV as explicit constraints. \addaccess{However, these differential equations in the optimization problem typically lead to computationally-expensive problems, due to the fine sampling needed in the discretization methods (shooting or collocation).}

The direct imposition of the dynamic equations can be avoided by leveraging the differential flatness map \diffFlatnessMap{}, which takes the acceleration $\mathbf{a}$ and $\psi$ and maps them to the rotation of the body frame. Due to the \emph{hedgehog theorem}\footnote{Also known as the \emph{hairy ball theorem} in the literature.} in $S^2$~\cite{bendixson1901courbes,brouwer1911abbildung}, this map is guaranteed to have at least one singularity when tried to be defined with a single continuous function. Several possible definitions of this differential flatness map are shown in Table~\ref{tab:singularities}, all of which \addaccess{satisfy~\eqref{eq:accel_rotation_constraint} by} construction. In the first two definitions, one body axis is obtained as the cross product of $\boldsymbol{b}_3\equiv\bar{\boldsymbol{\xi}}$ with a vector lying in the $xy$ world plane, and the remaining body axis is such that the resulting body frame is right-handed. These two definitions present a singularity whenever the normalized relative acceleration $\bar{\boldsymbol{\xi}}\in S^2$ is parallel to a vector defined by $\psi$ which lies in the $xy$ world plane. This means that, for a given $\psi$, the singularity appears for two $\bar{\boldsymbol{\xi}}$ that have a great-circle distance of $90^\circ$ with respect to the hovering condition. In aggressive flights, and due to numerical instabilities and artificial large changes of orientations near the singularity, this closeness between the hovering condition and the singularity can limit the set of possible accelerations for the planner.

The third definition of Table~\ref{tab:singularities} leverages the Hopf fibration~$\boldsymbol{h}(\cdot)$, which can be defined as a map $S^{3}\rightarrow S^{2}$~\cite{watterson2020control,lyons2003elementary}
that takes a unit quaternion $\boldsymbol{q}$ and produces the resulting
rotation of the vector $\arraycolsep=1.4pt\boldsymbol{e}_z:=\left[\begin{array}{cccc} 0 & 0 & 1 \end{array}\right]^{T}$ (see Fig.~\ref{fig:hopf}):
\[
\begin{bmatrix}0\\
	\boldsymbol{h}\left(\boldsymbol{q}\right)
\end{bmatrix}:=\boldsymbol{q}\circ\begin{bmatrix}0\\
	\boldsymbol{e}_{z}
\end{bmatrix}\circ\boldsymbol{q}^{-1}
\]
Making use now of the inverse image of the Hopf fibration, we have that $\boldsymbol{q}_{b}^{w}$ will be a composition of two rotations\footnote{Note that $\boldsymbol{q}_{\psi}$, $\boldsymbol{q}_{\boldsymbol{\xi}}$, and  $\boldsymbol{q}_{b}^w$ are guaranteed to be unit quaternions by construction.}: $\boldsymbol{q}_{\boldsymbol{\xi}}$, that aligns $\boldsymbol{b}_3$ with $\boldsymbol{\xi}$, followed by $\boldsymbol{q}_{\psi}$, which is a rotation around $\boldsymbol{\xi}$ by an angle $\psi$. 
Given a specific $\boldsymbol{\xi}$ (with $\bar{\boldsymbol{\xi}}\neq-\boldsymbol{e}_{z}$), the quaternion $\boldsymbol{q}_{b}^{w}=\boldsymbol{q}_{\boldsymbol{\xi}}\circ\boldsymbol{q}_{\psi}$ will then be a fiber (specifically a circle) in $S^{3}$ parametrized by $\psi$~\cite{lyons2003elementary}. 
The main advantage of the Hopf fibration over the previous two definitions is that the singularity only occurs when the UAV is inverted (i.e., when  $\bar{\boldsymbol{\xi}}=-\boldsymbol{e}_{z}$), which is the orientation that has the largest possible great-circle distance from the hovering configuration, and hence much less likely to happen.
\addaccess{Although the goal of this paper is not to plan highly aggressive trajectories, and hence any of the singularities shown in Table~\ref{tab:singularities} are unlikely to be reached, we use the Hopf map to automatically ensure the maximum distance to the singularity. 
Note also that, with the Hopf fibration, a} second chart could be used to cover the inversion point $\bar{\boldsymbol{\xi}}=-\boldsymbol{e}_{z}$, but the use of multiple charts, while computationally cheap in the controller level~\cite{watterson2020control} or in an intermediate check step in a decoupled $\mathbf{p}$--$\psi$ optimization~\cite{watterson2018geometric}, would significantly increase the computation time when embedded in the $\mathbf{p}$--$\psi$ joint planning optimization. This fact, together with the improbability of an upside-down configuration as being PA optimal, led us to the inclusion of only the first chart. 

Our work differs from other works that have used the Hopf fibration for UAVs \cite{watterson2020control,watterson2018geometric,welde2021dynamically} as follows: \addaccess{Ref.}~\cite{watterson2020control} uses the Hopf fibration only in a controller to track predefined trajectories. In~\cite{watterson2018geometric}, the Hopf fibration is used in the planner to find the charts in a step after the $\mathbf{p}$~optimization and before the $\psi$~optimization\addaccess{, and~\cite{welde2021dynamically} does not optimize~$\psi$}. We \addaccess{instead} propose to embed the Hopf fibration map directly on the $\mathbf{p}$--$\psi$ \emph{joint} optimization, as a way to directly obtain trajectories in $\text{SE}(3)$ that are dynamically feasible by construction, and with the crucial advantage of not needing to explicitly impose the dynamic equations as constraints in the optimization.

\subsubsection{Cost function} \label{subsec:cost_function}
A PA term in the objective function should maximize the presence in the FOV of the predicted position of the obstacle $i^*\in I$. However, this alone is not enough to guarantee good PA trajectories, since a fast moving projected obstacle in the image plane may cause significant blur, which can lead to stereo matching failure and consequently tracking failure. To take into account both the presence in the FOV and the blur, we use $\frac{\text{inFOV}(\cdot)}{\epsilon_1+\epsilon_2\left\Vert \dot{\boldsymbol{s}}\right\Vert ^{2}}$ as the running reward, where $\dot{\boldsymbol{s}}$ is the projected velocity in the image plane, %
\addrevision{and where $\epsilon_1$ and $\epsilon_2$ are nonnegative parameters such that $\epsilon_1+\epsilon_2>0$.}
Note how this reward is high if the predicted position of the obstacle is in the FOV with a small projected velocity, and it is approximately zero if the predicted position of the obstacle is not in the FOV, regardless of the value of the projected velocity. The position in the image plane of the projection of the obstacle can be obtained using the pinhole camera model as $\boldsymbol{s}(t):=\frac{f}{\left[\left(\tilde{\mathbf{p}}_{i^{*}}(t)\right)^{c}\right]_{z}}\begin{array}{c}
	\left[\left(\tilde{\mathbf{p}}_{i^{*}}(t)\right)^{c}\right]_{x:y}\end{array}$ (where each component of $\boldsymbol{s}$ is expressed in meters, not in pixels), and
$$\left(\tilde{\mathbf{p}}_{i^{*}}(t)\right)^{c}:=\boldsymbol{T}_{b}^{c}\boldsymbol{T}_{w}^{b}(t)\left(\tilde{\mathbf{p}}_{i^{*}}(t)\right)^{w}$$ 
\[
\boldsymbol{T}_{w}^{b}(t):=\left[\begin{array}{cc}
	\text{rot\ensuremath{\left(\boldsymbol{q}_{\boldsymbol{\xi}}(t)\circ\boldsymbol{q}_{\psi}(t)\right)}} & \mathbf{p}(t)\\
	\boldsymbol{0}^{T} & 1
\end{array}\right]^{-1}
\]
As detailed in Table \ref{tab:Notation}, the discontinuity of the function $\text{inFOV}(\cdot)$ is addressed by approximating it with a sigmoid function.

In addition to the PA term explained above, we also add two terms in the cost function to maximize the smoothness in position (by minimizing jerk) and $\psi$ (by minimizing $\ddot{\psi}$), and a terminal cost that penalizes the distance between $\mathbf{p}(t_{f})$ and $\boldsymbol{g}$.

\subsubsection{Collision avoidance and dynamic limits constraints}\label{subsubsec:collision_avoidance_dyn_lim}

For the obstacle avoidance of dynamic obstacles, we first create a polyhedral outer representation of both the trajectory of the agent and of the obstacle (see Fig. \ref{fig:explanation_trajs}): For the agent, we make use of the MINVO basis \cite{tordesillas2020minvo} (a polynomial basis that finds the simplex with minimum volume enclosing a polynomial curve) to obtain the set of control points $\left(\mathcal{Q}_{j}^{\text{MV}}\right)_{\text{agent}}$ whose convex hull encloses each segment $j$ of the agent. Similarly, for each obstacle~$i$, we first compute the MINVO control points of the segment~$j$ of the predicted mean $\left(\mathbf{p}_{i}\right)^{w}(t)$, and then we inflate it with $\text{norminv}(\delta)\cdot\mathbf{\boldsymbol{\sigma}}_{i}\left(t_{\text{end \ensuremath{j}}}\right)$,
half of the sides the AABB (axis-aligned bounding box) of the obstacle $i$ and half of the sides of the AABB of the agent. Here, $\delta \in[0,1]$ is the percentile of the standard normal distribution, and hence it encodes the desired level of conservativeness in the inflation. The resulting polyhedron is denoted as $\mathcal{C}_{ij}^{\text{MV}}$. 

To ensure safety between the agent and the obstacle $i$, we then impose linear separability constraints (via planes) between $\left (\mathcal{Q}_{j}^{\text{MV}}\right )_\text{agent}$ and $\mathcal{C}_{ij}^{\text{MV}}$. The separating planes are found during the initial guess search for the position spline \addaccess{(see section~\ref{subsec:init_guess_pos})}, and are held fixed in the optimization. The MINVO basis is used in a similar way to impose low-conservative constraints in the velocity space. In the acceleration and jerk spaces, the MINVO control points are the same as the B-Spline control points. \addaccess{These constraints on $\mathbf{v}$, $\mathbf{a}$, $\mathbf{j}$, and $\dot{\psi}$ serve as a conservative approximation of the real actuator constraints of the motors of the UAV, while allowing us to reduce the complexity of the optimization problem. }

\subsubsection{Optimization problem}\label{subsec:opt_problem_panther}

Including the initial state and the final hovering condition, the optimization problem is\footnote{Time dependence of the variables in the cost function has been omitted for simplicity.}:
\begin{empheq}[box=\fbox]{flalign*}
	&\underset{\mathbf{p}(t)\in\mathcal{S}_{3,m}^{3}, \psi(t)\in\mathcal{S}_{2,m}^{1}}{\boldsymbol{\min}}\alpha_{\mathbf{{j}}}\int_{t_{\text{in}}}^{t_{f}}\left\Vert \mathbf{j}\right\Vert ^{2}dt+\alpha_{\psi}\int_{t_{\text{in}}}^{t_{f}}\left(\ddot{\psi}\right)^{2}dt& \\
	&-\alpha_{\text{FOV}}\int_{t_{\text{in}}}^{t_{f}}\frac{\text{inFOV}(\boldsymbol{T}_{c}^{w},\left(\mathbf{p}_{i^*}\right)^{w})}{\epsilon_1+\epsilon_2\left\Vert \dot{\boldsymbol{s}}\right\Vert ^{2}}dt+\alpha_{\boldsymbol{g}}\left\Vert \mathbf{p}(t_{f})-\boldsymbol{g}\right\Vert^{2}\\
	&\text{s.t.} & \\ \quad
	& \qquad\mathbf{x}(t_{\text{in}})=\mathbf{x}_{\text{in}},\quad \mathbf{v}(t_{\text{f}})=\boldsymbol{0},\quad \mathbf{a}(t_{\text{f}})=\boldsymbol{0}\addaccess{,\quad \dot{\psi}(t_\text{f})=0}&\\
	& \qquad\boldsymbol{n}_{ij}^{T}\boldsymbol{q}+d_{ij}<0\quad\mspace{2mu}\forall\boldsymbol{q}\in\left(\mathcal{Q}_{j}^{\text{MV}}\right)_{\text{agent}},\;\forall i\in I,\;\forall j\in J&\\
	& \qquad\text{abs}\left(\boldsymbol{v}\right)\le\boldsymbol{v}_{\text{max}}\quad\mspace{6mu}\forall\boldsymbol{v}\in\left(\mathcal{V}_{j}^{\text{MV}}\right)_{\text{agent}},\;\forall j\in J&\\
	& \qquad\text{abs}\left(\boldsymbol{a}_{l}\right)\le\boldsymbol{a}_{\text{max}}\quad\forall l\in L_{\mathbf{p}}\backslash\{n_{\mathbf{p}}-1,n_{\mathbf{p}}\}&\\
	& \qquad\text{abs}\left(\boldsymbol{j}_{l}\right)\le\boldsymbol{j}_{\text{max}}\quad\;\forall l\in L_{\mathbf{p}}\backslash\{n_{\mathbf{p}}-2,n_{\mathbf{p}}-1,n_{\mathbf{p}}\}&\\
	& \qquad\text{abs}\left(\Psi_{l}\right)\le\Psi_{\text{max}}\mspace{15mu}\forall l\in L_{\psi}\backslash\{n_{\psi}\}
\end{empheq}

Here, \addaccess{$\arraycolsep=1.4pt\mathbf{x}:=\left[\begin{array}{ccccc}
	\mathbf{p}^{T} & \mathbf{v}^{T} & \mathbf{a}^{T} & \psi & \dot{\psi}\end{array}\right]^{T}$,
 $\left\{\alpha_{\mathbf{{j}}},\;\alpha_{\psi},\;\alpha_{\text{FOV}},\;\alpha_{\boldsymbol{g}}\right\}$}
 are nonnegative weights, and the decision variables are the control points of the splines $\mathbf{p}(t)$ and $\psi(t)$. The degrees chosen for the splines $\mathbf{p}(t)$ and $\psi(t)$ are, respectively, 3 and 2, which are a good trade-off between computation time and dynamic feasibility for a UAV~\cite{mellinger2011minimum}. \addrevision{The units of the weights are such that the corresponding term is dimensionless (see section~\ref{sec:results_and_discussion}). An empirical method to select these weight values is as follows: First set $\alpha_{\boldsymbol{g}}$  to a large value to ensure that the final location is near $\boldsymbol{g}$. Then, $\alpha_{\text{FOV}}$, together with $\epsilon_1$ and $\epsilon_2$, are tuned to obtain a good presence of the obstacle~$i^*$ in the FOV. Finally, $\alpha_{\mathbf{{j}}}$ and $\alpha_{\psi}$ are progressively increased to improve the smoothness of $\mathbf{p}(t)$ and $\psi(t)$, without significantly deteriorating the FOV cost.  }

To solve this optimization problem, we utilize the Interior Point Optimizer Ipopt~\cite{wachter2006implementation}\footnote{\addrevision{We classify an Ipopt solution as successful when Ipopt returns \texttt{Solve\_Succeeded} (locally optimal solution) or \texttt{Solved\_To\_Acceptable\_Level} (solution satisfying the acceptable tolerance level). For more details, see \cite{IpoptReturnCodes}.}}
interfaced through CasADi~\cite{Andersson2019} with \addaccess{MA27 and }MA57~\cite{hsl}
as the linear \addaccess{solvers} of Ipopt. \addrevision{All these optimization tools were installed and run onboard the UAV in the real-world experiments (section~\ref{sec:real_world_experiments}).} We approximate the PA term of the cost function using the composite Simpson's rule for numerical integration~\cite{simpson2020}. 

\section{Results and Discussion}\label{sec:results_and_discussion}

\newcommand{\noPA}{\textit{no PA}}
\newcommand{\PAdec}{\textit{PA dec}}
\newcommand{\Wang}{\textit{Wang}}
\newcommand{\psiSweep}{\textit{$\psi$ sweep}}

\definecolor{color_notinFOV_front}{RGB}{121,132,144}
\definecolor{color_notinFOV_behind}{RGB}{201,202,201}
\definecolor{color_inFOV}{RGB}{93,188,172}

\definecolor{color_noPA}{RGB}{1,113,188}
\definecolor{color_PAdec}{RGB}{216,82,24}
\definecolor{color_psisweep}{RGB}{236,176,31}
\definecolor{color_wang}{RGB}{125,46,141}
\definecolor{color_wang_light}{RGB}{202,121,219}
\definecolor{color_panther}{RGB}{119,172,48}
\begin{figure*}
	\sidesubfloat[]{\includegraphics[width=0.98\textwidth]{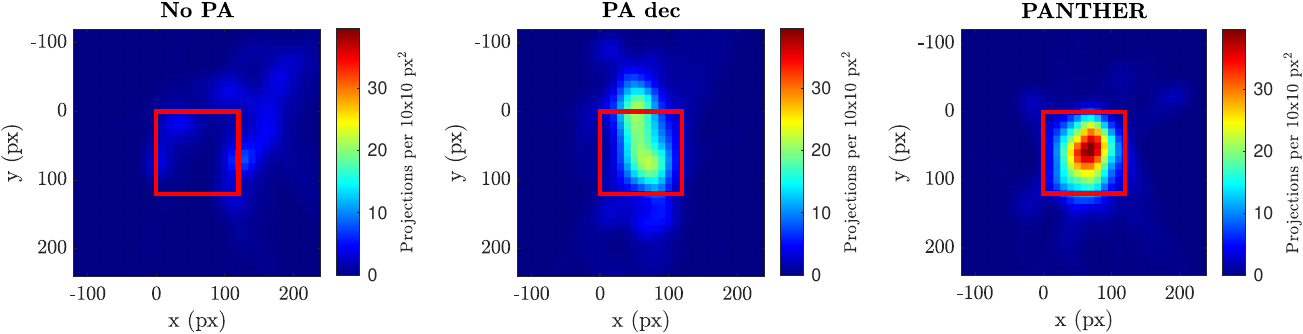}\label{fig:histogram_projections}}\quad
	\vskip 0.3cm
	\sidesubfloat[]{\includegraphics[width=0.45\textwidth]{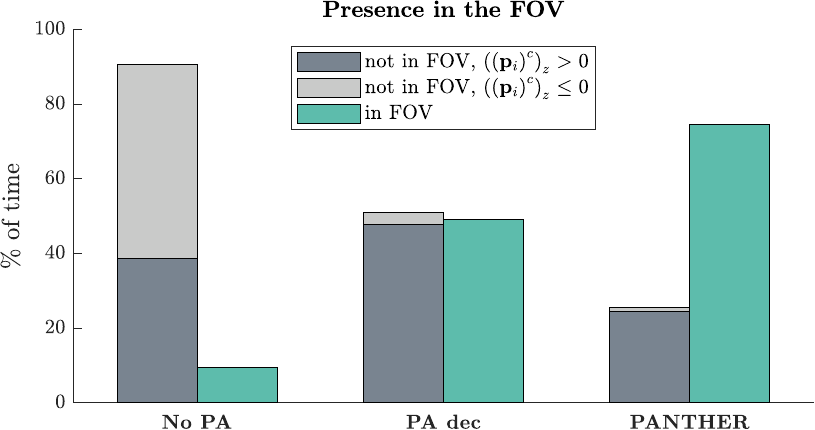}\label{fig:percentage_in_fov}}%
	\sidesubfloat[]{\includegraphics[width=0.24\textwidth]{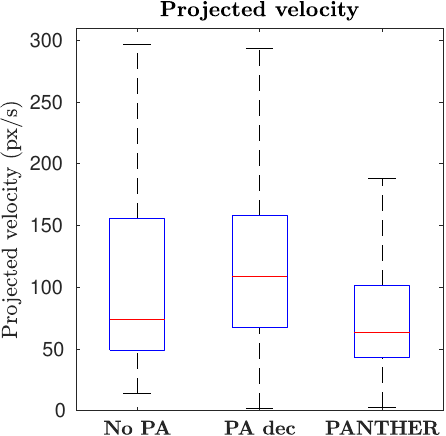}\label{fig:projected_vel}}%
	\sidesubfloat[]{\includegraphics[width=0.24\textwidth]{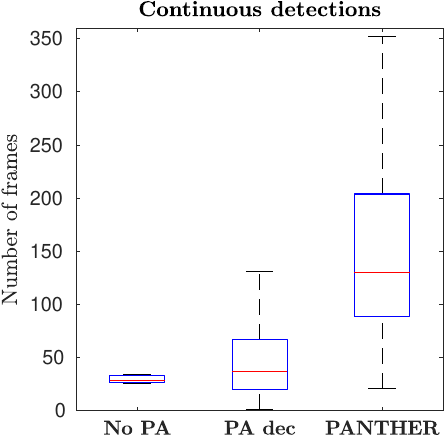}\label{fig:continouos_detections}}
	\vskip 0.3cm
	\sidesubfloat[]{\includegraphics[width=0.98\textwidth]{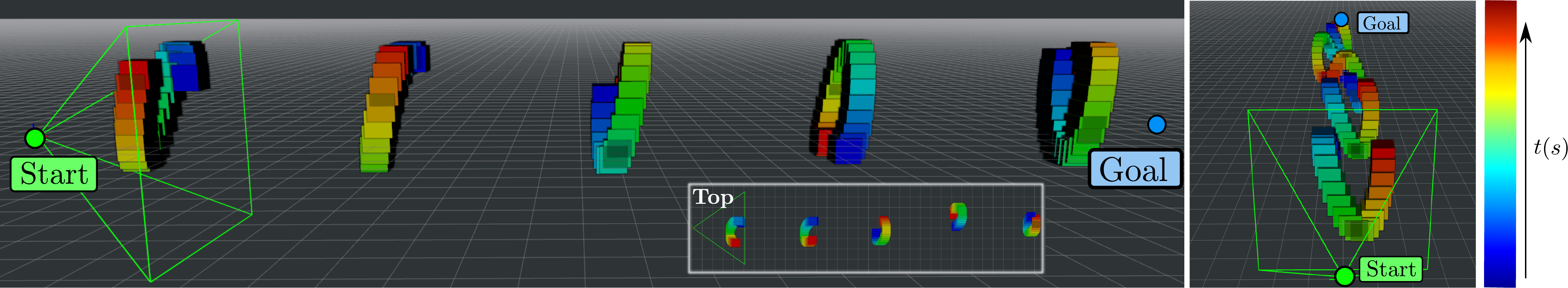}\label{fig:corridor_colored}}%
	\vskip 0.3cm
	\sidesubfloat[]{\includegraphics[width=0.98\textwidth]{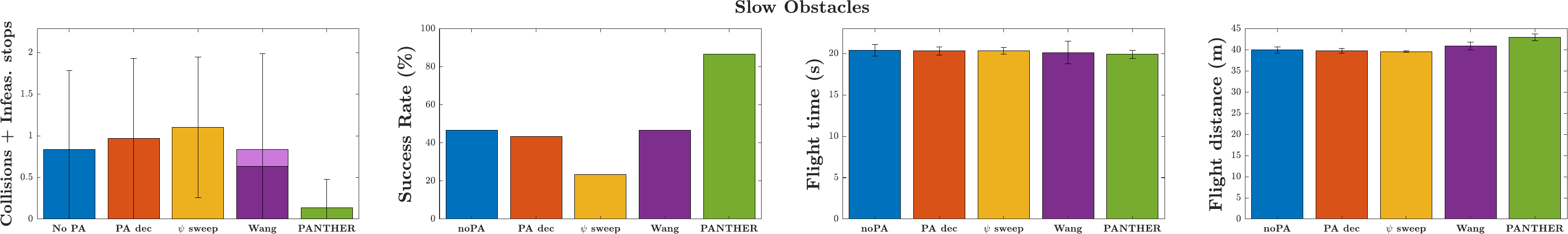}\label{fig:slow_simulation}}%
	\vskip 0.3cm
	\sidesubfloat[]{\includegraphics[width=0.98\textwidth]{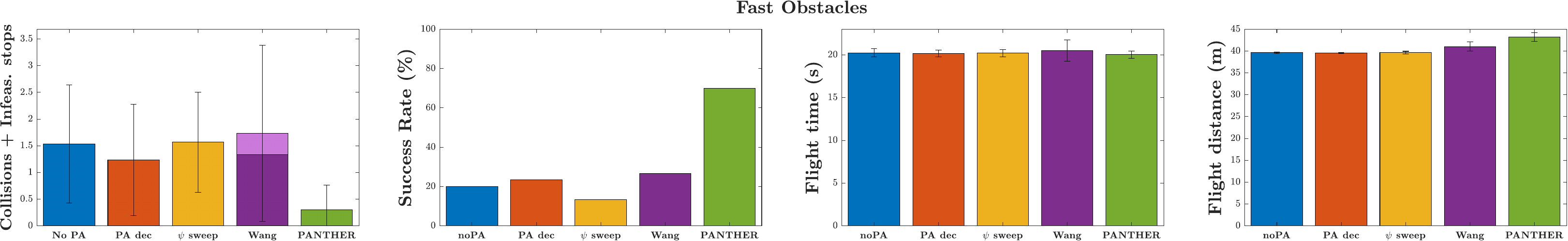}\label{fig:fast_simulation}}%
	\vskip 0.3cm
	\caption{ (\textbf{A}) Projections of the obstacle onto the image plane in the single-obstacle simulation experiments. The red square is the image
		plane, so any projection out of this region is not in the FOV of the
		camera. (\textbf{B}) Percentage of the time the obstacle was not in the FOV but in front of the camera (\tikzrectangle[black,fill=color_notinFOV_front]{10pt}), not in the FOV and behind the camera (\tikzrectangle[black,fill=color_notinFOV_behind]{10pt}), and in the FOV (\tikzrectangle[black,fill=color_inFOV]{10pt}). (\textbf{C}) Velocity of the projection of the centroid of the obstacle onto the image plane. Higher projected velocities produce larger blur in the image.  (\textbf{D}) Number of frames for each continuous detection. (\textbf{E}) Corridor simulation with five dynamic obstacles following random trefoil-knot trajectories~\cite{trefoil2020}. The green pyramid represents the FOV of the camera.  (\textbf{F, G}) Results for the corridor simulations with slow and fast obstacles, respectively. The algorithms considered are \noPA{} (\tikzrectangle[black,fill=color_noPA]{10pt}), \PAdec{} (\tikzrectangle[black,fill=color_PAdec]{10pt}), \psiSweep{} (\tikzrectangle[black,fill=color_psisweep]{10pt}), \Wang{}~\cite{wang2021autonomous} (\tikzrectangle[black,fill=color_wang]{10pt}), and PANTHER (\tikzrectangle[black,fill=color_panther]{10pt}). In the left plot of both subfigures, \tikzrectangle[black,fill=color_wang_light]{10pt} represents the number of infeasible stops of algorithm~\cite{wang2021autonomous}. The other algorithms have zero infeasible stops.   }%
\end{figure*}

\subsection{Simulation experiments}\label{sec:simulation_experiments}

All the simulation experiments are run in an  \texttt{AlienWare Aurora r8} desktop \addaccess{running Ubuntu 20.04 and} equipped with an Intel\textsuperscript{\textregistered} Core\textsuperscript{TM} i9-9900K CPU, 3.60GHz$\times$16 and 62.6~GiB. Moreover, and to focus the comparisons on the properties of the trajectories obtained by the planner, we assume, for all the algorithms benchmarked in simulation, that the UAV can perfectly track the trajectories obtained by the planner. 

\newcommand{\pantherweights}[9]{$c_{\Psi_{\text{max}}}=#1$, $c_{\text{FOV}}=#2$, $c_\psi=#3$~$\text{rad}^{-2}$, $\alpha_{\mathbf{{j}}}=#4$~${\text{s}^5}/{\text{m}^2}$, $\alpha_{\psi}=#5$~$\text{s}^3/\text{rad}^{2}$, $\alpha_{\text{FOV}}=#6$, $\alpha_{\boldsymbol{g}}=#7$~$\text{m}^{-2}$, $\epsilon_1=#8$, and $\epsilon_2=#9$~${\text{s}^2}/{\text{m}^2}$}

\subsubsection{Single obstacle}\label{sec:single_obstacle}
We first test PANTHER in an environment with a box-shaped obstacle of size $0.2\times0.2\times0.2$~m\textsuperscript{3} that follows a trefoil-knot~\cite{trefoil2020} trajectory. \addaccess{During 60 s, the UAV is commanded to continuously fly between two different locations whose centroid is the area where the obstacle is moving.}  
The camera has an image size of $120\times120\;\text{px}^{2}$, a limited FOV of $60^\circ \times 60^\circ$, and runs at a rate of $60$~Hz. \addrevision{The weights used for this simulation are \pantherweights{10^6}{1}{0}{10^{-6}}{0}{20}{70}{0.3}{0.45}.} To focus this comparison on the capabilities of the planner, we let the agent perfectly know the trajectory of the obstacle \addrevision{in} these simulations. We compare
the following three approaches:
\begin{enumerate}
	\item \textbf{No PA:} $\psi$ is held constant and only the
	smoothness in position and terminal goal costs are optimized. \addaccess{Works that do not plan $\psi$ include, e.g.,~\cite{chen2016online, falanga2020dynamic, sanket2020evdodgenet}.} 
	\item \textbf{PA with position and} $\psi$ \textbf{decoupled}: Translation
	$\mathbf{p}$ is optimized first (as in the method \noPA{}) and then it is held fixed while
	$\psi$ is optimized with the PA term. We will refer to this algorithm as \textbf{PA dec}. \addaccess{This decoupling is done in, e.g., \citePADec{}.}
	\item \textbf{PANTHER} (ours): Joint optimization of $\mathbf{p}$ and $\psi$. 
\end{enumerate}

As will be explained in \addaccess{section~\ref{subsec:cost_function}}, two important metrics that characterize a good PA trajectory are the presence of the obstacle in the FOV and the norm of the projected velocity, which quantifies the blur. The percentage of time the obstacle was in the FOV of the camera is shown in Fig.~\ref{fig:percentage_in_fov}. PANTHER is able to keep the obstacle inside the FOV \addaccess{7.9 and 1.5} times more than the algorithms \noPA{} and \PAdec{}, respectively. As \PAdec{} decouples position and $\psi$ in the optimization, the UAV lacks the ability to modify the
spatial path (only $\psi$) to generate a better overall trajectory.

To qualitatively show the area of the projection, we apply a Gaussian filter to the histogram of the projection of the centroid of the obstacle onto each $10\times10\;\text{px}^{2}$ cell of the image plane. The results are shown in Fig.~\ref{fig:histogram_projections}, where we can see that PANTHER is able to keep the obstacle inside the FOV limits much better, and \addrevision{more frequently}, than methods \noPA{} and \PAdec{}. 

The velocity of the projection of the centroid of the obstacle onto the image plane is shown in Fig.~\ref{fig:projected_vel}, which highlights that PANTHER is able to obtain a~\addaccess{$18$\% and~$34$\%} decrease in the mean of the norm of the projected velocity with respect to \noPA{} and \PAdec{}, respectively, achieving, therefore, a much less blurred projection of the obstacle than those two methods.

Finally, and as a continuous detection of the dynamic obstacle is crucial to achieve a good tracking and prediction, we show in Fig.~\ref{fig:continouos_detections} the boxplot of the number of frames of each continuous detection for the different algorithms. A continuous detection is defined as a set of consecutive frames for which the obstacle stayed in the FOV of the camera. On average, PANTHER is able to achieve continuous detections of \addaccess{$155$}~frames, while the mean number of frames per continuous detection for methods \noPA{} and \PAdec{} are \addaccess{$39$}~and \addaccess{$46$}~frames, respectively. 

\subsubsection{Several obstacles}\label{subsec:simulation_several_obstacles}

We now test PANTHER in a simulation with several obstacles. The environment consists of a corridor of length of $39$~m along the $x$ direction with five dynamic obstacles that move following random trefoil-knot trajectories~\cite{trefoil2020}, see Fig.~\ref{fig:corridor_colored}. In all these simulations, the agent only has access to the size, current position, and velocity of the obstacles that are inside the FOV of the camera. The FOV of the camera is $70^\circ \times 70^\circ$, and has a sensing range of 5~m. The dynamic limits are $\boldsymbol{v}_\text{max}=2.6\cdot\boldsymbol{1}$~\velunits{}, $\boldsymbol{a}_\text{max}=15.5\cdot\boldsymbol{1}$~\accelunits{}, $\boldsymbol{j}_\text{max}=\addaccess{50.0}\cdot\boldsymbol{1}$~\jerkunits{}, and $\Psi_\text{max}=\pi$~\addrevision{rad/s.} \addrevision{The weights used for PANTHER in these simulations are \pantherweights{10^6}{1}{0}{10^{-7}}{0}{40}{25}{0.3}{10^{-5}}.}  The UAV is constrained to remain in $y\in[-4,4]$~m and $z\in[-4,4]$~m at all times.

For the benchmark, we use the algorithms explained before (\noPA{}, \PAdec{}, and PANTHER), and the two additional algorithms:

\begin{itemize}
	\item Algorithm~\cite{wang2021autonomous}, proposed by Wang et al. This approach is not perception aware, but $\psi$ tries to make the FOV of the camera point to the direction of travel. We will refer to this algorithm as \textbf{Wang}. Note also that this algorithm does not have constraints on $\boldsymbol{j}_\text{max}$ and that it has a different $\psi$ convention (it uses definition~1 of Table \ref{tab:singularities}). %
	\item $\psi$ \textbf{sweep}: $\psi$ follows a sinusoidal trajectory that varies in $[-90^\circ, 90^\circ]$ as follows:
	$$\psi(t)=\frac{\pi}{2}\sin\left(\frac{\Psi_\text{max}}{\pi/2} t\right)$$
\end{itemize}

\begin{figure*}
	\begin{centering}
		\addrevisiongraphics{\includegraphics[width=1\columnwidth]{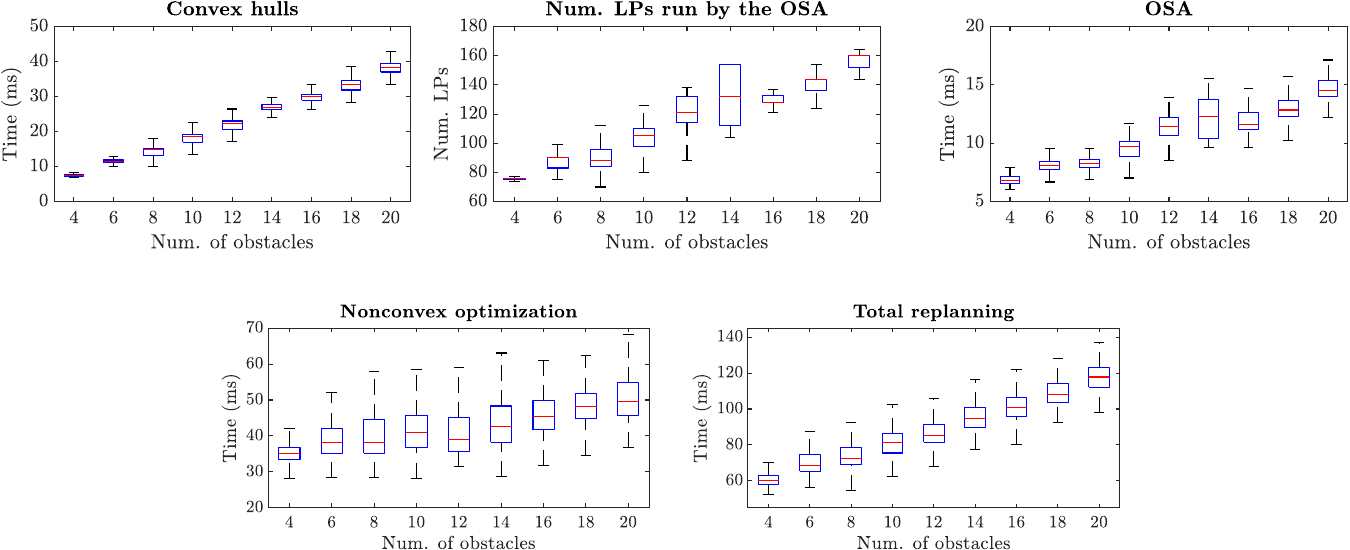}}
		\par\end{centering}
	\caption{\addrevision{Computational analysis of different parts of the replanning step of PANTHER as a function of the number of obstacles. From left to right, and top to bottom: computation time of the generation of the convex hulls, number of linear programs (LPs) run by the OSA, computation time of the OSA, computation time of the nonconvex optimization, and total replanning time.  \label{fig:comp_times_simulation}}}
\end{figure*}

We test two scenarios with different maximum velocities of the obstacles. In the slow scenario, the obstacles move with velocities up to \addaccess{2.12}~\velunits{}, while in the fast scenario, the obstacles move with velocities up to \addaccess{4.07}~\velunits{}. In the results, we compare the number of collisions, infeasible stops, success rate, flight time, and flight distance. An infeasible stop happens when the drone passes instantly from a nonstop condition ($\mathbf{v}\neq\boldsymbol{0}$ or $\mathbf{a}\neq\boldsymbol{0}$) to a stop condition ($\mathbf{v}=\boldsymbol{0}$ and $\mathbf{a}=\boldsymbol{0}$). A run is considered successful if the UAV is able to reach the end of the corridor while not colliding with any of the obstacles. \addaccess{To make these simulations closer to real-world applications, where no prior information about the trajectories of the obstacles may be available, a simple constant velocity model is used in the predictor. The obstacles themselves are moving along trefoil-knot trajectories~\cite{trefoil2020}}. 

The results, for 30 different runs per algorithm, are shown in Figs.~\ref{fig:slow_simulation} and~\ref{fig:fast_simulation} for the slow and fast environments, respectively.  In the slow scenario, PANTHER is able to succeed \addaccess{$87\%$} of the runs, while the other algorithms have a success rate \addrevision{below} \addaccess{$47\%$}. None of the algorithms present infeasible stops except \Wang{}, that has a mean of \addaccess{0.2} infeasible stops per run (light purple in Fig.~\ref{fig:slow_simulation}). In the fast scenario, PANTHER succeeds \addaccess{$70\%$} of the runs, while all the other algorithms have a success rate \addrevision{below} \addaccess{$27\%$}. In terms of flight times and flight distances, most of the algorithms achieve very similar results in both scenarios, with a total flight time of \addrevision{approximately} $20$~s, and an \addrevision{approximate} total flight distance of $41$~m. \addaccess{The total flight distance for PANTHER is \addrevision{approximately} $3$~m more than the rest of the algorithms. This is expected, because PANTHER has the ability to modify the spatial path to maximize the visibility of the obstacles. Even with this longer flight distance, the flight time of PANTHER is very similar (and sometimes even \addrevision{shorter}) than the rest of the algorithms.}

\begin{figure*}
	\sidesubfloat[]{\includegraphics[width=0.98\textwidth]{imgs/frozen_all.pdf}\label{fig:frozen_all}}\quad
	\vskip 0.5cm
	\sidesubfloat[]{\includegraphics[width=0.98\textwidth]{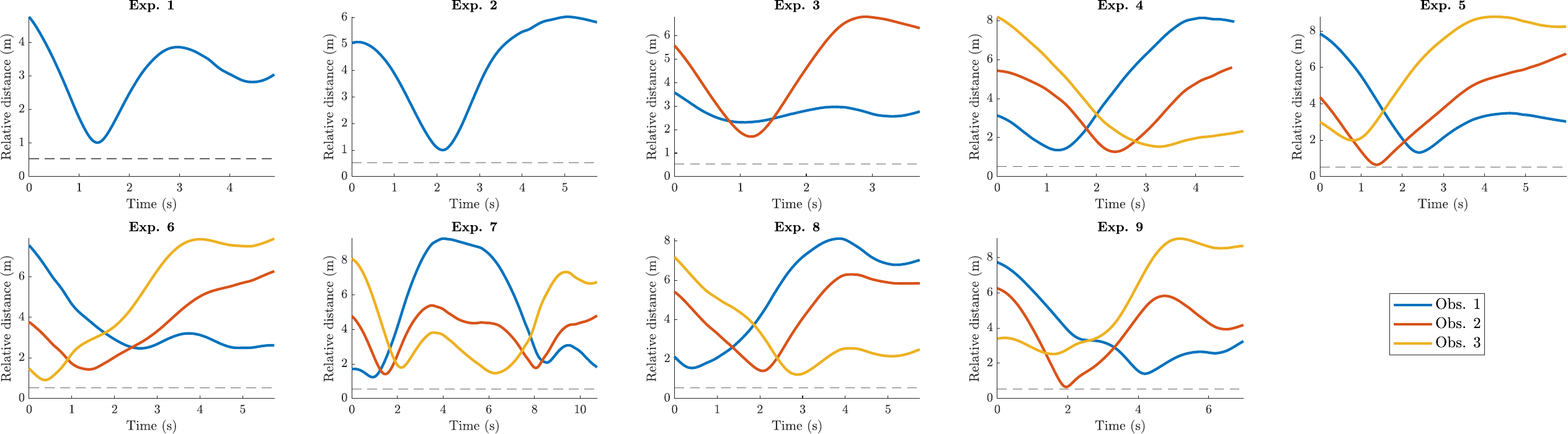}\label{fig:rel_distances_hw}}%
	\vskip 0.2cm
	\caption{ (\textbf{A}) Composite images of all the nine experiments. For visualization purposes, only the second half of Experiment~7 is shown. The table below every image shows the number of obstacles, flight distance, maximum velocity, maximum relative velocity (with respect to the obstacles), and flight time of each experiment. The number of obstacles is one, two, and three for the experiments 1-2, 3, and 4-9 respectively. (\textbf{B}) Relative distances between the agent and each one of the obstacles. Any relative distance above the dashed line guarantees safety. }%
\end{figure*}

\begin{figure*}
	\textbf{Experiment 3} \vskip -0.3cm
	\sidesubfloat[]{\includegraphics[width=0.98\textwidth]{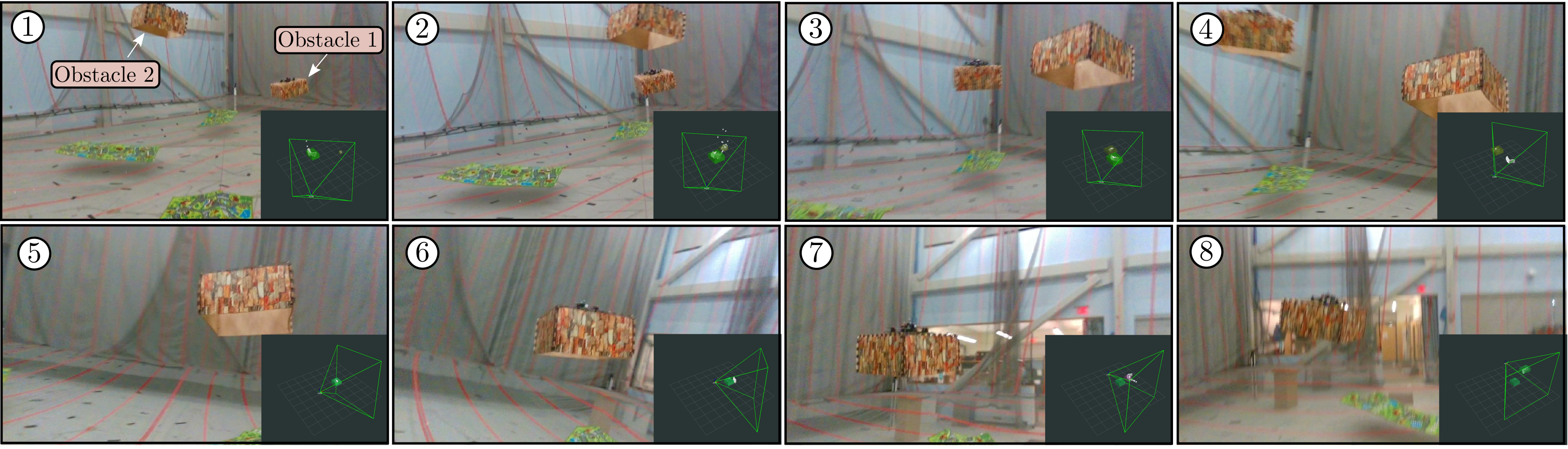}\label{fig:onboard_camera_hwexp_3_enhanced}}\quad
	\vskip 0.5cm
	\textbf{Experiment 6} \vskip -0.3cm
	\sidesubfloat[]{\includegraphics[width=0.98\textwidth]{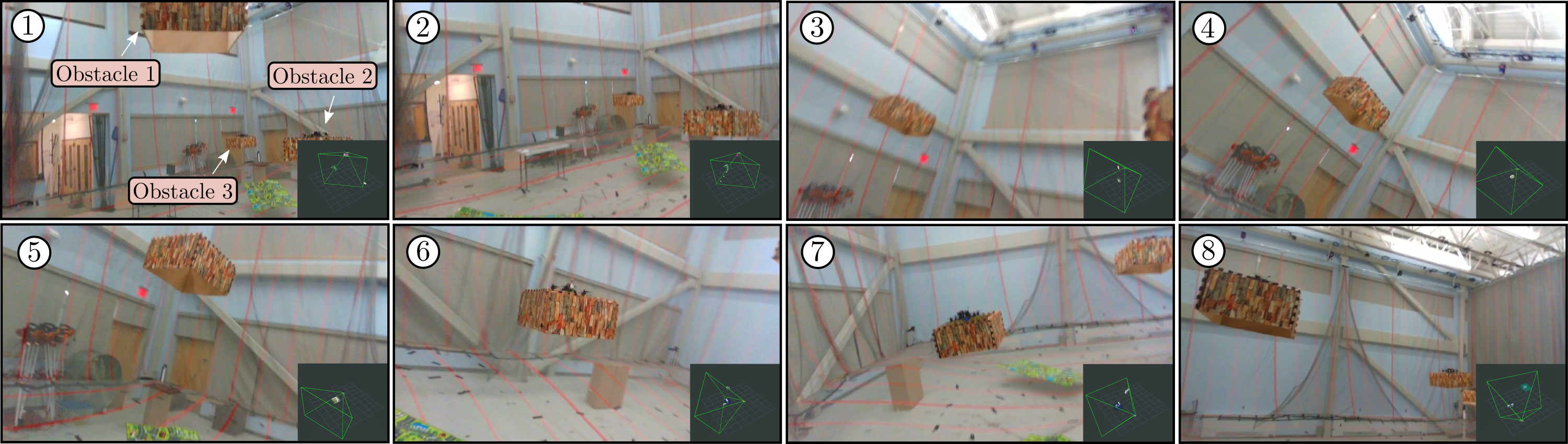}\label{fig:onboard_camera_hwexp_6_enhanced}}%
	\vskip 0.5cm
	\textbf{Experiment 9} \vskip -0.3cm
	\sidesubfloat[]{\includegraphics[width=0.98\textwidth]{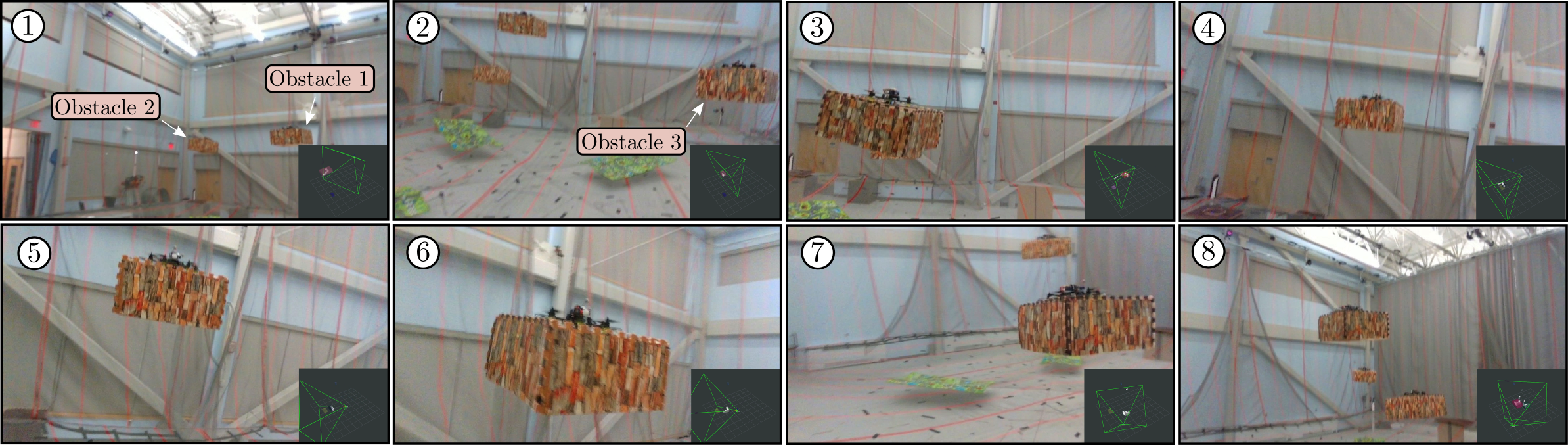}\label{fig:onboard_camera_hwexp_9_enhanced}}%
	\vskip 0.5cm
	\sidesubfloat[]{\includegraphics[width=0.98\textwidth]{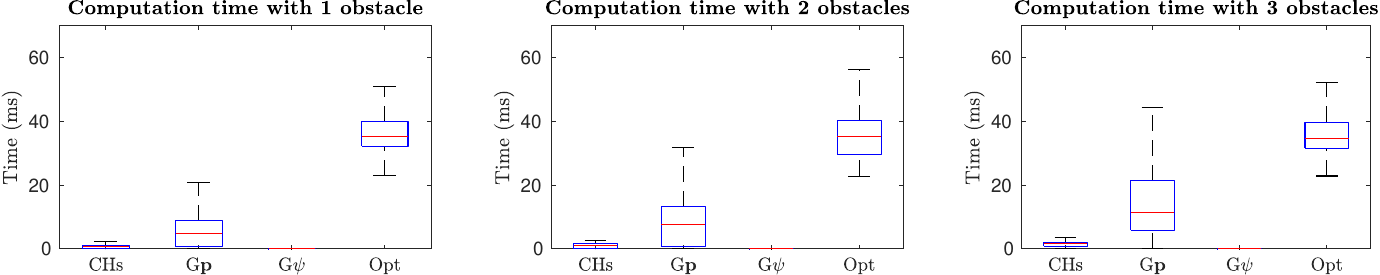}\label{fig:comp_times}}%
	\caption{ Snapshots of the onboard camera in experiments~3~(\textbf{A}), 6~(\textbf{B}), and 9~(\textbf{C}). (\textbf{D}) Computation times for each part of a replanning step, measured on the onboard Intel\textsuperscript{\textregistered} NUC i7DNK. The tracker, predictor, and the depth camera were also running on this computer at the same time these times
		were measured. The notation used is: CHs (convex hull computation for the polyhedral outer representations), G$\mathbf{p}$ (generation of the planes and the guess for the position $\mathbf{p}(t)$),
		G$\psi$ (generation of the guess for $\psi(t)$) and Opt (Optimization
		time). }\label{fig:test}
\end{figure*}

\subsubsection{\addrevision{Computational analysis of the replanning step as a function of the number of obstacles}}\label{sec:complexity_analysis}

\addrevision{We now compare the computational cost of different parts of the replanning step of PANTHER. As the computational cost of each part highly depends on the specific position of the obstacles relative to the UAV, we perform a Monte Carlo analysis by randomly deploying obstacles (which follow trefoil-knot trajectories) in the spherical shell~\cite{sshell2021} limited by two spheres of radii $2$~m and $5$~m. The starting location $\arraycolsep=1.4pt\left[\begin{array}{ccc} 0 & 0 & 1\end{array}\right]^T$~m and $\boldsymbol{g}_{\text{term}}=\arraycolsep=1.4pt\left[\begin{array}{ccc} 6 & 0 & 1\end{array}\right]^T$~m are held fixed for every replanning iteration. The number of obstacles tested are $\{4,\;6,\;\hdots,\;18,\;20\}$, and, for each number of obstacles, we run 10 simulations of $5.0$~s each. For these simulations, the UAV includes all the deployed obstacles in the planning problem (i.e., the set $I$ contains the indexes of all the obstacles deployed), and we let the UAV know the trajectory of the obstacles perfectly. The weights used are the same as the ones used in section~\ref{subsec:simulation_several_obstacles}. The results are shown in Fig.~\ref{fig:comp_times_simulation}, where can see that the computation time required for the convex hull generation, the OSA, the optimization, and the total replanning time change approximately linearly with the number of obstacles. Similarly, the number of linear programs run by the OSA also changes approximately linearly with the number of obstacles. 
The average solve time of one of these linear programs is $0.09$~ms. 
}

\addrevision{To obtain the~$\psi$ initial guess (section~\ref{subsec:init_guess_psi}), the average runtime of the Dijkstra's algorithm on the~$\psi$ graph is $0.137$~ms, and the average runtime to fit a spline to the $\psi$ samples (Eq.~\ref{eq:fit_yaw_samples}) is $0.048$~ms.} 

\addrevision{These results above show the computational analysis for the different parts of the replanning step of PANTHER (convex hull computation, generation of the $\mathbf{p}(t)$ and $\psi(t)$ initial guesses, and nonconvex optimization). For the computational cost of the tracker and predictor using real point clouds, see section~\ref{sec:real_world_experiments}. The well-known results regarding the complexity analysis of the Hungarian algorithm are given in~\cite{kuhn1955hungarian, edmonds1972theoretical}.}

\subsection{\addaccess{Real-world} experiments}\label{sec:real_world_experiments}

We run an extensive set of hardware experiments, where a UAV needs to go from a starting point to a goal location while avoiding unknown dynamic obstacles. The UAV used is equipped with a Qualcomm\textsuperscript{\textregistered} SnapDragon Flight, an Intel\textsuperscript{\textregistered} NUC i7DNK, and an Intel\textsuperscript{\textregistered} RealSense Depth camera D435i. The tracker, planner, and the camera run on the Intel\textsuperscript{\textregistered} NUC, while the control and state estimation run on the Qualcomm\textsuperscript{\textregistered} SnapDragon Flight. \addrevision{Note that the main onboard computer (Intel\textsuperscript{\textregistered} NUC) has similar computational power to the onboard hardware used in the recent literature (e.g.,~\cite{zhou2020raptor,  chen2016online, sanket2020evdodgenet, wang2021autonomous})}. \addrevision{For the controller, we run the approach presented in \cite{lopez2016low,watterson2020control} at 100~Hz to generate the desired orientation and angular rates from $\mathbf{p}(t)$ and $\psi(t)$. The commanded thrusts for the motors are then found from these attitude commands using a geometric controller~\cite{lee2010geometric}, which is run at IMU rate (500~Hz). }For state estimation, \addrevision{we use a} visual inertial odometry \addrevision{(VIO)} package~\cite{sdVIO} \addrevision{running at 30~Hz that leverages an extended Kalman filter to fuse} the IMU measurements of the SnapDragon and the images of its downward-facing camera. \addrevision{To obtain a high-rate state estimate, we then integrate forward the IMU (which runs at 500~Hz) between consecutive VIO estimates. } 

The IMU of the RealSense camera is not used. All the computation of this UAV is running onboard, and it does not have any prior knowledge of the trajectories and specific shape/size of the obstacles. \addrevision{The weights used for these experiments are \pantherweights{10^6}{1}{0}{0.05}{0.1}{1}{2\cdot10^4}{0.1}{1}.}

To generate the dynamic obstacles, we use three other UAVs with a Qualcomm\textsuperscript{\textregistered} SnapDragon Flight, and equip
them with a box-shaped frame of $\approx0.6\times0.6\times0.3$~m\textsuperscript{3}.
The obstacles are following trefoil-knot trajectories~\cite{trefoil2020}.

A total of 9 experiments were performed \addrevision{(see attached video)}. The composite images of the trajectories flown by the agent and by the
obstacles, together with the number of obstacles, distance flown,
maximum velocity, maximum relative velocity with respect to the obstacles, and total flight time of each one of the experiments are shown in Fig.~\ref{fig:frozen_all}. Experiments 1 and 2 were done with one obstacle, experiment 3 with
two obstacles, and experiments 4-9 with three obstacles. The maximum
velocity achieved by the agent, 5.77 m/s, happened in experiment 7.
In that same experiment, the maximum relative velocity (6.28~m/s) with respect to the obstacles
is also achieved. The relative distances between the UAV and the obstacles are shown in Fig.~\ref{fig:rel_distances_hw}. Any relative distance above the dashed horizontal line guarantees safety between the agent and the corresponding obstacle. For experiments 3, 6, and 9, different snapshots of the onboard camera are shown in Figs. \ref{fig:onboard_camera_hwexp_3_enhanced}, \ref{fig:onboard_camera_hwexp_6_enhanced}, and  \ref{fig:onboard_camera_hwexp_9_enhanced}, respectively. Note how the planned trajectories try to keep an obstacle in the FOV at all times to aid in obstacle tracking and prediction.

The computation times are shown in Fig. \ref{fig:comp_times}.
All these computation times \addrevision{were} measured onboard, with the UAV
flying, and with the depth camera node and the tracker running on the same
computer (Intel\textsuperscript{\textregistered} NUC i7DNK). The mean total replanning times are 48.70, 51.66, and 58.59 ms for the experiments with 1, 2, and 3 obstacles respectively. The point cloud of the camera is generated at $90$ Hz, and the tracker (clustering, assignment, and prediction) is able to process each point cloud in $\approx 8.6$ ms.

\section{Conclusion}\label{sec:conclusions}

\addaccess{This work derived PANTHER, a perception-aware (PA) trajectory planner in dynamic environments. PANTHER is	able to couple together the translation and the full rotation in the optimization, leading to PA trajectories computed in real time that maximize the presence of the obstacles in the FOV while minimizing their projected velocity. Extensive hardware	experiments in unknown dynamic environments, with all the computation running onboard, and with relative velocities of up to 6.3 m/s have shown its effectiveness.}

Our approach has also some limitations. Specifically, in the hardware experiments we observed the importance of the choice of the obstacle to include in the optimization (i.e., the choice of $i^{*}$, see Table~\ref{tab:Notation} and \addaccess{section~\ref{sec:several_obstacles}}): when should the UAV include a specific (already tracked) obstacle in the PA term of the optimization, in order to predict its trajectory more accurately to be able to avoid it, and when should the UAV turn around to explore unknown space? This highlights the trade-off between exploration and exploitation: too much focus on exploitation may lead to collision with obstacles that were never detected, and too much focus on exploration may lead to a very poor trajectory prediction, and hence to a collision as well. Optimally solving this trade-off is a promising direction for future work.

\addaccess{Another \addrevision{possible} direction of future work is to solve the trade-off between visibility and time optimality. This would entail adding the time minimization in the optimization problem of section~\ref{subsec:opt_problem_panther}, and would also allow to highlight the advantages of the Hopf fibration when flying aggressive trajectories that pass close to the singularity produced by the commonly-used maps presented in~\cite{mellinger2011minimum, faessler2017differential} (first two definitions of Table~\ref{tab:singularities}).}

\addrevision{Finally, another interesting research direction is how to incorporate disturbances in the planning problem, while still guaranteeing that the tracking error of the UAV remains bounded~\cite{lopez2019dynamic}. The incorporation of such disturbance information is especially important when flying outdoors under windy conditions, since a large deviation between the planned trajectory and the actual trajectory can provoke a collision with the obstacles. 
}

\section*{Acknowledgment}
The authors would like to thank Parker Lusk, \addrevision{Dr.}\ Kris Frey, \addrevision{Dr.}\ Kaveh Fathian, Yulun Tian, and Stewart Jamieson for helpful insights and discussions. \addrevision{The authors would also like to thank the anonymous reviewers, whose valuable feedback helped to improve the article.}

\bibliography{my_bib}
\bibliographystyle{IEEEtran}

\begin{IEEEbiography}[{\includegraphics[width=1in,height=1.25in,clip,keepaspectratio]{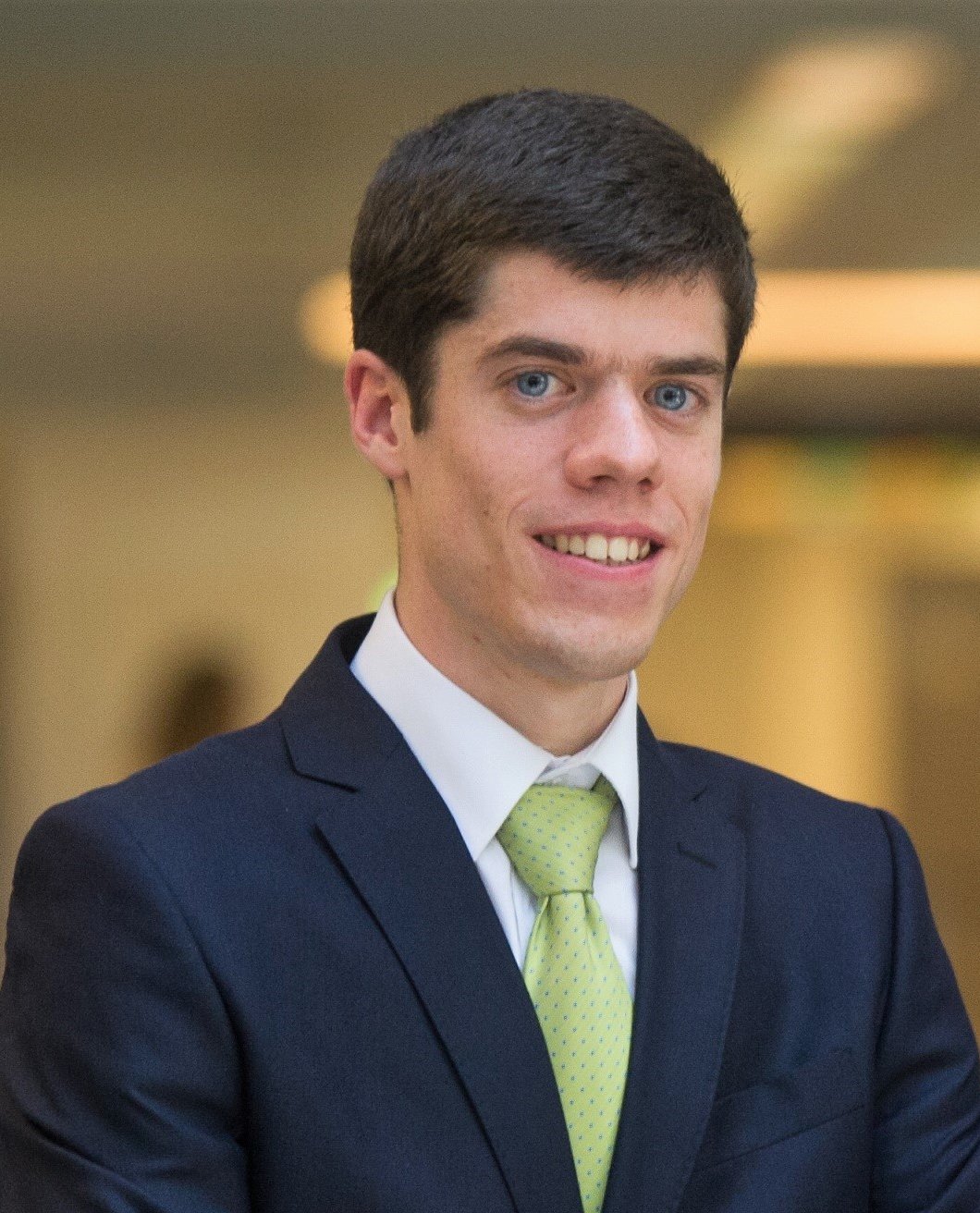}}]{Jesus Tordesillas}
	(Student Member, IEEE)
	received the B.S. and M.S. degrees in Electronic engineering and Robotics from the Technical University of
	Madrid (Spain) in 2016 and 2018 respectively. He then received his M.S. in Aeronautics and Astronautics from MIT in 2019. He is currently pursuing the Ph.D. degree with the Aeronautics and
	Astronautics Department, as a member of the Aerospace Controls Laboratory (MIT) under the supervision of Jonathan P. How. 
	His research interests include path planning for UAVs in unknown environments and optimization. His work was a finalist for the Best Paper Award on Search and Rescue Robotics in IROS 2019.
\end{IEEEbiography}

\begin{IEEEbiography}[{\includegraphics[width=2.9in,height=1.25in,clip,keepaspectratio]{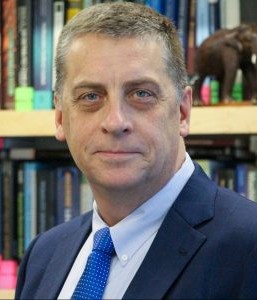}}]{Jonathan P. How}
	(Fellow, IEEE) received the
	B.A.Sc. degree from the University of Toronto (1987), and the S.M. and Ph.D. degrees in aeronautics and astronautics from MIT (1990 and 1993). Prior to joining MIT in 2000,
	he was an Assistant Professor at Stanford University. He is currently the
	Richard C. Maclaurin Professor of aeronautics and astronautics at MIT. Some of his awards include the IEEE CSS Distinguished Member Award (2020), AIAA Intelligent Systems Award (2020), 
		IROS Best Paper Award on Cognitive Robotics (2019), and the AIAA Best
		Paper in Conference Awards (2011, 2012, 2013). 
		He was the Editor-in-chief of IEEE Control Systems Magazine (2015--2019), is a Fellow of AIAA, and 
		was elected to the National Academy of Engineering in 2021.
\end{IEEEbiography}

\EOD

\end{document}